\documentclass{article} 
\usepackage[final]{colm2025_conference}
\usepackage{microtype}
\usepackage{hyperref}
\usepackage{booktabs}
\usepackage{xurl}     
\usepackage{hyperref} 
\usepackage{amsmath}
\usepackage{xcolor} 
\usepackage{lineno}
\usepackage{booktabs}
\usepackage{algorithm}
\usepackage{algpseudocode}
\usepackage{multirow}
\usepackage{longtable}
\usepackage{expex}
\usepackage{graphicx}

\lingset{glspace=.5em, everygla=\it, belowglpreambleskip=0ex, aboveglftskip=0ex} 

\definecolor{darkblue}{rgb}{0.1, 0.3, 0.8}
\hypersetup{colorlinks=true, citecolor=darkblue, linkcolor=darkblue, urlcolor=darkblue}

\title{The Gold Medals in an Empty Room: Diagnosing Metalinguistic Reasoning in LLMs with Camlang}

\author{
Fenghua Liu$^{1}$\quad Yulong Chen$^{1}$\quad Yixuan Liu$^{2}$\quad  Zhujun Jin$^{1}$\quad Solomon Tsai$^{1}$\\ \textbf{ Ming Zhong$^{3}$ }
  \\
  $^1$ University of Cambridge \quad $^2$ University of Oxford\quad $^3$ UIUC\\
  \centering{\texttt{\href{mailto:fl457@cam.ac.uk}{fl457@cam.ac.uk}}\quad\texttt{\href{mailto:yc632@cam.ac.uk}{yc632@cam.ac.uk}}}
  }

\begin{document}

\ifcolmsubmission
\linenumbers
\fi

\maketitle

\begin{abstract}
Large Language Models (LLMs) achieve gold-medal performance across many benchmarks, yet it remains unclear whether such success reflects genuine reasoning or pattern matching.
From a cognitive science perspective, an informative test is whether models can master an unfamiliar language through explicit metalinguistic deductive learning, a paradigm where human learners can reliably internalise grammatical systems through metalinguistic reasoning.
We address this question with {Camlang}, a novel constructed language that exhibits naturalistic yet unattested feature combinations.
Camlang consists of two explicit resources, a grammar book and a bilingual dictionary, which mirror adult second-language learning via explicit grammar rules and lexical lookup, and enable us to disentangle errors in morpho-syntax, lexical semantics, and sentence-level reasoning.
Human experiments show that these resources are sufficient for participants to acquire Camlang and successfully solve Camlang tasks.
To operationalise evaluation, we adapt CommonsenseQA into Camlang, creating {Camlang-CSQA-v0}, the first task in a broader suite where solving questions requires applying grammar rules and lexical mappings.
Experimental results show that GPT-5 achieves 98\% EM accuracy in English but only 47\% in Camlang, far below human performance at 87\%, while other state-of-the-art reasoning LLMs perform even worse. 
Human verification further reveals that most model successes stem from shallow lexical alignment while GPT-5 shows emerging metalinguistic awareness to a limited extent but not systematic grammatical mastery as humans. 
Camlang establishes a cognitively grounded evaluation paradigm that exposes fundamental gaps between current models and human metalinguistic competence.


{\color[rgb]{1,0.2,0.5} 
\textbf{Note:} This ArXiv paper presents ongoing work. The current experimental results are preliminary, based on our existing data. We may further revise Camlang and will extend the tasks in our benchmark, perform additional experiments, and update the analysis in future versions.}

\end{abstract}

\section{Introduction}

Large Language Models (LLMs) have performed remarkably in language tasks, demonstrating versatility and adaptability~\citep{grattafiori2024llama, openai2023gpt4,openai2024gpt4o,deepseekai2025deepseekr1,openai2025gpt5}. 
Yet, a fundamental question arising from this observed performance remains unresolved: are these models truly competent in understanding human language, or are they merely performing statistical pattern matching from their massive training corpora? 
This debate is central to both AI and cognitive science, and has sparked discussion on whether modern LLMs possess human-like understanding and reasoning~\citep{chomsky2023nytimes,hinton2024nobel}.
Validating such claims requires an evaluation setting where success cannot be gauged exclusively through memory or recalling patterns from training data, but instead depends on the ability to acquire and apply linguistic competence through an understanding of the underlying systems which constitute natural languages.

One natural way to test this ability is to present LLMs in a language they have never encountered before, mirroring the explicit rule–based learning scenarios found in human second language (L2) acquisition~\citep{klein1986second,ELLIS2005209}.
Cognitive science research shows that adult learners can successfully acquire grammatical rules and vocabulary of a new language and apply them to novel cases when they use their native language as an explanatory medium~\citep{lichtman2012child,rebuschat2012implicit}.\footnote{Note this is opposed to infant bi-/multi-lingual acquisition~\citep{williams2016implicit,hummel2021introducing}, which typically relies on natural, implicit linguistic stimulus from the environment.}
Such success builds on their domain-general reasoning and metalinguistic skills, which enable learners to systematically internalise the explicitly presented rules and map the new vocabulary onto pre-existing knowledge in the mental lexicon~\citep{gombert1992metalinguistic,bialystok2001metalinguistic,nagy2007metalinguistic}.
Controlled experiments in psycholinguistics and neuroscience further show that human adults can acquire unfamiliar language systems from grammar rules, and subsequently integrate these rules with their existing semantic and reasoning abilities~\citep{reber1967implicit,kim1997distinct,musso2003broca}. 
This leads to the central question of this paper: can LLMs learn and use an unseen language when given only explicit grammar rules and vocabulary like human learners? 

Existing LLM evaluation protocols are ill-suited to investigate this question.
Most LLM benchmarks are built for English~\citep{wangglue,wang2019superglue,talmor2019commonsenseqa, chen2024see} and thus inevitably share real-world linguistic distributions with pre-training data~\citep{balloccu2024leak}, making it difficult to distinguish the application of linguistic knowledge and reasoning from pattern matching.
Even evaluations on extremely low-resource or endangered languages are prone to data leakage~\citep{artetxe-etal-2020-cross,chen-etal-2023-revisiting,zhang2024teaching}, due to the fact that their grammar books, dictionaries, and sample texts are often well-documented and publicly available online.\footnote{Such as Kalamang: \url{https://langsci-press.org/catalog/book/344}.}
While a potential alternative is to train an LLM exclusively on one language (e.g., English) and test it on another (e.g., Mandarin Chinese) to ensure novelty in the tested language~\citep{chirkova2024zero}, this is impractical for modern LLM systems.
Most state-of-the-art LLMs are inherently multilingual~\citep{katz2024gpt,deepseekai2025deepseekr1}, and training a fully monolingual model solely for this evaluation would be prohibitively costly.
It would also limit the model's general capabilities, making the results less informative about the potential of leading systems~\citep{qorib-etal-2025-just}.

In this work, we introduce Camlang, a typologically novel constructed language.
Camlang is constructed through a principled design process that combines typologically attested features in a novel configuration and encodes them in two explicit resources: a grammar book and a bilingual dictionary.
The grammar book specifies phonological, morphological, and syntactic rules, while the dictionary provides a lexicon generated and expanded via controlled mechanisms (root generation, derivation, compounding, and loans), ensuring that the language is both plausible and learnable for humans and computational models.
Camlang offers three key advantages.
First, the novelty ensures that evaluation cannot be reduced to recalling seen distributions, but instead tests whether models can generalise linguistic competence to a genuinely unfamiliar system.
Second, the grammar–lexicon separation enables fine-grained diagnosis of model behaviour, making it possible to distinguish failures of structural parsing from those of semantic interpretation or lexical lookup.
Furthermore, human learnability guarantees that the benchmark reflects cognitively realistic language acquisition, providing a meaningful point of comparison between human learners and models.
Together, these properties make Camlang a cognitively grounded and leakage-free benchmark for testing whether LLMs can go beyond statistical pattern recognition and engage in explicit deductive language learning.

Using the Camlang grammar book and English–Camlang dictionary, we construct a small test set, Camlang-CSQA-v0, by manually translating questions and answers from CommonsenseQA~\citep{talmor2019commonsenseqa} into Camlang. 
Since LLMs already excel at CommonsenseQA in English~\citep{krause2023commonsense,davis2023benchmarks}, the bottleneck of Camlang-CSQA-v0 shifts from commonsense reasoning to applying explicit grammar rules and lexical mappings.
The resulting dataset allows us to isolate the ability to integrate newly learned grammatical knowledge with pre-existing world knowledge in a controlled setting, directly paralleling explicit second-language learning in human adults, and serves as a pilot benchmark for scaling Camlang to broader evaluation tasks.

Our experiments compare multiple state-of-the-art reasoning LLMs, including GPT-5~\citep{openai2025gpt5} and DeepSeek-R1~\citep{deepseekai2025deepseekr1}, with a human participant. 
Results reveal a sharp contrast: while models achieve 85--98\% EM accuracy on the English version, their performance drops sharply to 21--47\% on Camlang. 
The human participant reaches near-native accuracy (87\%) using only the grammar and dictionary, which also demonstrates the practical learnability of Camlang. 
This suggests that, unlike human learners who naturally integrate dictionary lookup into their reasoning pipeline, current LLMs struggle to operationalise external resources when faced with Camlang's morphological complexity, often defaulting to internal pattern matching. 
Human verification further reveals an important nuance that GPT-5's reasoning traces are more often incomplete rather than incorrect, indicating a degree of metalinguistic awareness and partial rule-following ability, while it falls short of systematic grammatical mastery. 
In contrast, other LLMs, such as DeepSeek-R1 and GPT-o3~\citep{openai2024gpt3o}, show little evidence of such awareness, with most of their successes attributable to shallow lexical alignment. 
Together, these findings establish Camlang as both a reliable diagnostic resource and a controlled benchmark for probing reasoning in unfamiliar languages.

In summary, the contribution of this paper includes: 
\begin{itemize}
    \item It introduces a typologically plausible yet novel language, equipped with a grammar book and dictionary.
    \item We provide cognitive grounding by empirically validating Camlan's learnability through human experiments.
    \item By adapting CommonsenseQA into Camlang, we test not only grammatical rule acquisition but also the integration of explicit rules with commonsense reasoning.
    \item We complement standard EM accuracy with human verification of reasoning traces, exposing the reliance of LLMs on shallow alignment rather than genuine metalinguistic deductive reasoning.
\end{itemize}

Beyond Camlang-CSQA-v0, we are currently extending Camlang to a broader suite of tasks (e.g., translation, parsing, and mathematical reasoning) and are exploiting its explicit grammar and dictionary for fine-grained diagnostic analyses, such as error localisation, rule generalisation, and learning-curve studies, ultimately establishing Camlang as a versatile testbed for probing the limits of LLM reasoning and their convergence with, or divergence from, human-like linguistic intelligence.

\section{Related Work}
\subsection{Natural Language Evaluation and Benchmarks}

Early benchmarks such as GLUE \citep{wangglue}, SuperGLUE \citep{wang2019superglue}, and CommonsenseQA \citep{talmor2019commonsenseqa} focus on general natural language understanding (NLU) and knowledge reasoning. 
While initially challenging, these benchmarks have been soon surpassed at human-level performance by LLMs~\citep{chen-etal-2023-unisumm, hu2024improving, openai2024gpt4o,wang2025hitchhiker}, motivating the development of more challenging benchmarks that target multi-step reasoning and specialised knowledge \citep{hendryckstest2021mmlu,srivastava2023beyond,patel-etal-2024-multi,gao2024omnimath,chen2024see}. 
Subsequent evaluations focused on tasks that are challenging even for human experts, yet where LLMs have demonstrated remarkable success, such as the U.S.\ Bar Exam~\citep{katz2024gpt}, USMLE~\citep{kung2023performance}, MATH~\citep{hendryckstest2021mmlu}, AIME~\citep{aime}, and even International Mathematical Olympiad problems.

While these results are impressive, they are all conducted in familiar natural languages and established task formats, allowing LLMs to exploit potential memorisation or pattern matching from training data. 
In contrast, our work employs a controlled constructed language to eliminate prior exposure, enabling a cleaner assessment of a model's ability to learn and apply entirely new linguistic and reasoning rules from scratch.

\subsection{Artificial, Constructed, and Programming Languages in NLP}\label{sec:Conlang}

Constructed language (Conlang) in cognitive science has long been used to study how humans acquire grammatical structure from limited exposure \citep{reber1967implicit,marcus1999rule}.  
Inspired by these paradigms, NLP researchers have proposed controlled synthetic benchmarks such as SCAN \citep{lake2018generalization}, NACS \citep{bastings2018jump}, and gSCAN \citep{ruis2020benchmark} to evaluate compositional generalisation of NLP systems.  
More recently, \citet{white2021examining} and \citet{hopkins2022towards} use artificial languages to probe the inductive biases of neural language models and to improve ecological validity, respectively.
\citet{kallini-etal-2024-mission} extend this line of work by creating impossible languages via systematic violations of linguistic universals, demonstrating that LLMs learn naturalistic languages far more efficiently than impossible ones.
In contrast, our {Camlang} is deliberately designed to be typologically plausible yet absent from attested natural languages, ensuring novelty without introducing unnatural structures.  
This allows us to test whether LLMs can integrate explicitly presented grammatical rules with pre-existing semantic and commonsense reasoning to solve downstream tasks, moving beyond pure grammatical learnability to functional language use.

Beyond artificial grammars, programming languages have also been used as testbeds for evaluating LLMs' rule-following and compositional abilities \citep{chen2021evaluating,li2022competition}. While such languages provide a fully formal and unambiguous setting, they differ fundamentally from natural language: programming languages are intentionally engineered for executability and determinism, lacking paraphrastic variation, pragmatic inference, and the typological diversity of human languages~\citep{clark1996using,allamanis2018survey,chomsky2014aspects}. 
In contrast, our {Camlang} is designed to be typologically plausible and naturalistic while still controlled, enabling us to probe whether LLMs can integrate explicit grammatical instruction with semantic and commonsense reasoning in a way that is closer to natural language learning.

Most importantly, the emphasis on naturalness distinguishes our approach from the above systems. 
If a constructed language were based on unattested or linguistically implausible rules~\citep{kallini-etal-2024-mission}, the evaluation would risk collapsing into a test of symbol manipulation rather than language learning, producing results that are uninterpretable for both cognitive and NLP perspectives. 
By ensuring that Camlang's grammatical features are all attested in human languages and that their combinations respect known typological constraints, we guarantee that it remains within the distribution of humanly possible languages. 
This design choice makes our evaluation cognitively comparable to adult second-language learning and scientifically credible as evidence of whether LLMs can acquire and apply genuine linguistic competence from explicit description.

\subsection{Cross-lingual Evaluation on Low-resource Languages}

Evaluation of LLMs on low-resource languages has relied on multilingual benchmarks such as XQuAD \citep{artetxe-etal-2020-cross}, TyDi QA \citep{clark2020tydi}, and FLORES \citep{guzman-etal-2019-flores}, as well as task-specific datasets in certain low-resource languages~\citep{adebara2022towards, chen-etal-2023-revisiting, zhang-etal-2025-read}.
While these datasets cover a broad range of typologically diverse languages, many are derived from publicly available sources such as Wikipedia, parallel corpora, or well-documented analyses and dictionaries.  
This leads to the data leakage problem that example sentences may have been seen during model pre-training \citep{carlini2023extracting,dong2023abilities}, allowing models to recall rather than genuinely learn the target language.  

Recent work has turned to controlled settings based on Olympiad-in-Linguistics-style puzzles and unseen-language tasks \citep{chi2024modeling,goyal2025iolbench,zhang2024teaching,lian2025lingbench++}, which construct problems from scratch to test grammatical rule induction from minimal examples.  
However, these puzzles often focus on isolated phenomena without providing a coherent, system-wide grammar.  
In contrast, our {Camlang} offers a fully specified, typologically plausible yet unattested language with a systematic grammar and lexicon, enabling controlled and repeatable evaluation of whether LLMs can learn and consistently apply explicit grammatical rules, and integrate them with pre-existing commonsense knowledge in downstream reasoning tasks.

\subsection{Human Adults Language Learning and Metalinguistic Skills}
Second-language acquisition research shows that adults are capable of learning grammatical rules through explicit instruction \citep{krashen1982principles,ELLIS2005209}.
Metalinguistic awareness—the ability to reflect on and manipulate linguistic form—has been shown to facilitate grammar learning~\citep{tunmer1984development,thomas1992metalinguistic,nagy1995metalinguistic}, particularly among bilinguals~\citep{bialystok2001bilingualism}.
Psycholinguistic studies further demonstrate that learners can acquire novel morphosyntactic systems from written descriptions alone, and then apply these rules in new contexts by integrating them with existing semantic and reasoning abilities.
In NLP, recent work has examined whether LLMs can similarly make use of explicit linguistic descriptions.
For example, prior work has shown that prompting LLMs to generate grammatical explanations~\citep{scheinberg2025explain} or providing them with grammar books and dictionaries~\citep{behzad-etal-2024-ask,hus2024back,zhang2024teaching} can improve performance on grammar judgment and low-resource translation tasks.

Compared to them, our work greatly differs in two ways. 
First, instead of relying on existing natural languages, where partial exposure through pre-training cannot be excluded, we use {Camlang}, a fully constructed, typologically plausible yet unattested language, ensuring zero data contamination.
Second, our grammar book is complete and systematically designed, covering multiple levels of linguistic structure and enabling targeted assessment of whether LLMs can internalise and consistently apply explicit grammatical rules.
By integrating this with commonsense reasoning tasks, we create a controlled setting to test whether LLMs can combine newly learned linguistic rules with pre-existing world knowledge for functional language use.

\section{The {Camlang} Language}\label{sec:camlang_language}

{Camlang} is a constructed language designed to evaluate whether LLMs can understand and apply new linguistic knowledge from explicit rules, rather than from prior pre-trained knowledge. 
In this work, the term ``\textit{constructing the language}'' refers specifically to the creation of two resources: {a grammar book}, written in English with {Camlang} samples, that systematically describes the phonology, morphology, syntax, and other aspects of the language, and {an English–{Camlang} bilingual dictionary}, covering all relevant lexical stems used in the evaluation tasks.\footnote{Unless otherwise specified, in the later sections, the terms \textit{grammar book} and \textit{dictionary} refer to the {Camlang} resources we create.}
Note that we describe {Camlang} as an \textit{unseen language} not because each of its individual linguistic features is unattested in other natural languages, but because the combination of features in {Camlang} is, to the best of our knowledge,\footnote{We acknowledge the possibility that a language similar to {Camlang} may exist or have existed, but has not been documented in the linguistic record to our knowledge.} absent from any known language. 

The subsequent subsections provide a detailed description of the Camlang language and its construction. 
We first introduce the design principles of {Camlang} in \S~\ref{sec:design principles}.
Then, we detail our construction and implementation in \S~\ref{sec:construction}.
In \S~\ref{sec:overview_camlang}, we present a high-level linguistic description of the language, focusing on its major phonological, morphological, and syntactic characteristics.
Finally, in \S~\ref{sec:camlang_analysis}, we provide an empirical evaluation of {Camlang} in a broader context, demonstrating its clarity, learnability and novelty as a naturalistic system while confirming that it is distinct from other known languages.

\subsection{Design Principles and Philosophies}\label{sec:design principles}
As stated, our goal is to simulate adult language learning for a new language in order to evaluate the reasoning abilities of LLMs. 
To serve this purpose, the language must reasonably resemble genuine human languages rather than arbitrary symbol systems~\citep{kallini-etal-2024-mission}, which makes naturalness a primary consideration. 
If the grammar were constructed with arbitrary, unattested rules, the evaluation would reduce to a symbol-manipulation game rather than a meaningful simulation of language learning. 
In such cases, both humans and models can fail simply because the rules are unnatural or even unlearnable, making results uninterpretable. 
At the same time, {Camlang} should be sufficiently rich to cover diverse grammatical phenomena, novel enough to avoid coinciding with attested languages, and compositional so that its rules apply systematically and productively. 
Thus, the development of {Camlang} is driven by the below principles:
\begin{itemize}
    \item Naturalness, to ensure plausibility as a human language;
    \item Richness, to cover a wide spectrum of grammatical phenomena;
    \item Novelty, to avoid the overlaps of the overall combination of linguistic features with attested languages;
    \item Compositionality, to guarantee systematic and productive rule application.
\end{itemize}
Together, these principles provide both a practical foundation and criteria.

\paragraph{Naturalness.}
Naturalness requires that the constructed language could, in principle, exist as a human language on Earth, and thus must fall within the distribution of humanly possible languages. 
We further distinguish three types of criteria for naturalness: 
(1) naturalness involves fundamental properties of human language, such as design features \citep{hockett1960origin} and the capacity for recursion \citep{doi:10.1126/science.298.5598.1569}, and;
(2) naturalness requires that the grammatical features employed are attested in human languages. 
Unattested and empirically unproven innovations~\citep{kallini-etal-2024-mission}, such as a phonological rule turning every second phoneme into ``/\textit{a}/'' if the word begins with ``/\textit{b}/'', should be excluded, and;
(3) naturalness concerns the probabilities of feature combinations, where certain features as rarely co-occurring and others as mutually implicative~\citep{greenberg1963some}. 
These linguistic universals~\citep{greenberg1963some, gorriereduplication, moravcsik1978distribution} constitute the most complex constraints that {Camlang} must respect to remain naturalistic.

\paragraph{Richness.}
We aim to make the language rich enough for our purposes~\citep{haspelmath2005world,evans2009myth}. In particular, (1) it should possess a diverse range of grammatical features, allowing us to analyse which features are more challenging for LLMs and what kinds of linguistic reasoning are lacking in LLMs, and;
(2) it should possess enough vocabulary and grammatical constructions to cover basic language usages, and;
(3) it should allow one meaning to be expressed in alternative stylistically differentiated ways.
Also, it should be noted that {Camlang} is not intended to match the full richness and complexity of a natural language such as English.
Instead, it is a controlled, moderately complex system designed to probe specific aspects of LLM language learning.

\paragraph{Novelty.}
The constructed language is intended to be a novel language, with unseen lexicon and grammar. 
Therefore, it is important to limit coincidences and overlaps with other languages that we know to our knowledge~\citep{comrie1989language,evans2009myth}. 
In particular, we consider two aspects: (1) from the lexicon aspect, the linguistic forms (spellings, in this case) should avoid coinciding with linguistic forms of other languages that mean the same thing, and; 
(2) from the grammar aspect, it should also be different from those of other languages to a significant extent. 
We emphasise again that the grammatical features are all attested in human languages, so the novelty refers to novel combinations of features, which means that the underlying logic of this language should differ from other languages.

\paragraph{Compositionality.}
The grammar of the constructed language is intended to be learnable with explicit description and demonstrative examples only.
Therefore, it must be compositional, in the sense that grammar rules are factorised and composed productively~\citep{chomsky2014aspects}: 
any word stem of a given part-of-speech (PoS) can take the specified inflectional slots, and functional categories always co-occur according to fixed linear orders and scope relations, unless exceptional rules or items, which should remain peripheral, are specifically mentioned in the grammar book. 
This ensures that the legality and linearisation of unseen morpheme combinations are decidable from the grammar book alone.


\paragraph{Other Considerations.}
Apart from the above four main principles, the design of the constructed language must also be regular, concise, and straightforward enough to ensure tractability so that both humans and computational models can process it reliably and efficiently.
Meanwhile, we deliberately abstract away from peripheral world-building considerations, such as putative geography~\citep{janda2013geography}, sociocultural embedding~\citep{lantolf1995sociocultural}, or franchise-specific lore~\citep{androic2019dialects} (e.g., Tolkienian settings or Parseltongue), and treat the constructed language purely as a typologically plausible testbed for explicit-rule learning.
It should also be noted that the main considerations sometimes lead to contradictions. 
For example, a more naturalistic feature combination is more common and therefore less novel, and a richer grammar with more features is less naturalistic due to the greater number of uncommon features and marked feature combinations. 
This requires the four considerations to be balanced in a trade-off in our implementation.


\subsection{Construction and Implementation}\label{sec:construction}

As discussed in \S~\ref{sec:Conlang}, we have not seen any attempt to construct a full-fledged plausibly human language for research purposes. 
Conlanging, or the construction of new full-fledged languages, is practised mostly for personal recreation, interlanguage communication (Auxlangs such as Esperanto~\citep{janton1993esperanto} and Interlingua~\citep{lu2018neuralinterlinguamultilingualmachine}), or world-building in fiction, most often without the need for rigorous scientific methodology. 
Therefore, the construction and implementation of {Camlang} as a research tool is mostly exploratory and pioneering to our knowledge.

Unlike existing conlangs~\citep{lake2018generalization,bastings2018jump}, Camlang is designed with {explicit operational goals}: it must be both learnable by humans and usable by computational models. 
To this end, Camlang is grounded in two resources: a grammar book and a dictionary. 
The grammar book specifies how meanings are compositionally expressed through phonology, morphology, and syntax, while the dictionary documents the lexical inventory and its distribution. 
Together, these artefacts provide the operational foundation that enables both humans and models to learn and use the language.  
The design of Camlang's grammar and dictionary is primarily led by a Cambridge linguistics student with prior experience in conlanging. 
Another three linguistics students from Cambridge and Oxford serve as core contributors in the refinement process. 
This ensures that all features are systematically analysed from multiple perspectives.

We next describe the conception and organisation of the initial grammar book (\S~\ref{sec:grammar_book}) and dictionary (\S~\ref{sec:dictionary}).
We then explain how these resources evolves through iterative human translation exercises and revisions (\S~\ref{sec:iterative_construction}), resulting in the final versions used in our experiments.

\subsubsection{Grammar Book}\label{sec:grammar_book}

The grammar book is the central artefact of {Camlang}, specifying how meanings are compositionally expressed through phonology, morphology, and syntax. 
We first describe how the characteristic features of {Camlang} are chosen and detailed, and then the organisation of the grammar book.

\paragraph{Feature Selection.} 
The naturalness and novelty principles dictate that {Camlang}'s features must all be attested but combined in an innovative configuration. 
With this goal, we start by creating the defining features of {Camlang} via the selection of existing features from language groups with distinct typologies, including but not limited to: vowel harmony from Turkic languages~\citep{korn1969types}, consonant mutation from Celtic languages~\citep{green2006independence}, and an active left periphery from Romance languages~\citep{beninca2004left}. 
To achieve richness and compositionality goals most directly, the author designed {Camlang} to have highly regular, rich morphology with prevalent simple exponence.

\paragraph{Specification.}
With the features preliminarily decided, we translate them into explicit phonological, morphological, and syntactic rules. 
We deliberately merge morphology and syntax in the grammar book into a single component (``Morphosyntax''), because many syntactic functions in {Camlang} are morphologically realised, and their description requires treating the two levels jointly. 
\begin{itemize}
    \item \textbf{Phonology Specification.} 
    In line with the novelty principle, the phonological traits borrowed from attested languages are systematically reinterpreted within {Camlang}'s original phoneme inventory and orthography, rather than copied directly. 
    For instance, both voiced and voiceless stops undergo aspirate mutation into voiceless aspirated stops—a pattern distinct from Middle Welsh, where only voiceless stops undergo aspirate mutation while voiced stops remain unaffected.
    \item \textbf{Morphosyntax Specification.} 
    To achieve richness and compositionality, we first identify core grammatical categories such as nominal number, verbal aspect, and clause types. 
    We then associate these categories with plausible morphological and syntactic patterns, including prefixation, suffixation, and verb fronting, and assign functional morphemes phonological shapes consistent with the designed phonology. 
    For instance, \textit{yes–no} questions are encoded both by verb fronting and by the attachment of a proclitic (``\textit{nAs=}'') containing a harmonising vowel underspecified for backness, in accordance with vowel harmony.
\end{itemize}

\paragraph{Organisation.} 
Language is conventionally described in a modular sequence, proceeding from phonetics, phonology, morphology, and syntax to semantics and pragmatics, gradually moving from linguistic form to linguistic meaning \citep{fromkin2017introduction}. 
In our case, phonetics is first excluded, since {Camlang} is not intended to be spoken. 
Accordingly, the initial grammar book is organised as:  

\begin{itemize}
    \item \textbf{Orthography and Phonology}, reflecting the rich phonological system and rules that drastically alter the written form;  
    \item \textbf{Nominal Morphology}, \textbf{Verbal Morphology}, and \textbf{Derivational Morphology}, capturing the highly productive morphology of {Camlang};  
    \item \textbf{Morphosyntax}, which subsumes syntax due to the strong interactions between syntactic structures and morphological marking.  
\end{itemize}

It should be noted that semantics and pragmatics are left largely unspecified in the initial version, both because our evaluation focuses on formal structure and because the dictionary captures much of the semantic content. 

As about to be discussed in \S~\ref{sec:iterative_construction}, the grammar book has been incrementally changed during refinement.

\subsubsection{{Dictionary}}\label{sec:dictionary}

The dictionary records the forms and meanings of {Camlang} lexemes and multi-word expressions, and thus serves as an indispensable reference for both humans and models. Guided by our design principles, we pursue two objectives: 
(1) a plausible, naturalistic distribution of lexical items in form and origin; and 
(2) minimal arbitrariness to support human and machine learnability.

\paragraph{Lexical Sourcing and Formation.}
To meet these objectives, we balance five sourcing mechanisms.
\begin{itemize}
    \item \textbf{Generated Roots (Native Camlang Word).} Given the phoneme inventory and phonotactics, we implement a root generator that samples mono- and disyllabic forms from a slot-conditioned probability table (see Appendix~\S~\ref{sec:root_generator}). 
    We select a newly generated form whenever a novel root is required, discarding accidental duplicates unless deliberate homonymy is desired. 
    More basic concepts tend to receive monosyllabic roots, more complex ones disyllabic roots (mimicking synchronically opaque derivations). 
    Sound symbolism occasionally informs choices, e.g., high front vowels for sharp objects vs.\ low round vowels for round objects \citep{cwiek2022bouba}.
    \item \textbf{Derived Words.} We exploit derivational morphology~\citep{tyler1989acquisition} to limit the number of independent roots. 
    Many derivatives are synchronically transparent and compositional, e.g., `\textit{cak}' (`\textit{to work}') $+$ `\textit{-mA\textsuperscript{4}}' (deverbal nominaliser) $\to$ `\textit{cakma}' (`\textit{work}'); others are conventionalised and semantically opaque, e.g., `\textit{kök}' (`\textit{to look}') $+$ `\textit{-GA\textsuperscript{4}s}' (abstract nominaliser) $\to$ `\textit{kökkys}' (`\textit{hope}'), both attested cross-linguistically.%
    \footnote{We use ``\textit{transparent/opaque}'' in the synchronic sense; c.f., standard discussions of derivational semantics~\citep{beard2017derivation}.}
    Derivation also includes calques (loan translations), e.g., `\textit{soruk}' (`\textit{compute}') $+$ `\textit{pI\textsuperscript{4}-}' (`\textit{agent}') $\to$ `\textit{pusóruk}' `\textit{computer}' \citep{shuttleworth2014dictionary}.
    \item \textbf{Compounds.} For semantically decomposable complex concepts~\citep{fff14662-cc84-3012-8104-515862aabe38}, we form compounds from existing words, thereby limiting new roots and maintaining agglutinative consistency, e.g., `\textit{kityb-chog}' (`\textit{desk}') from `\textit{kityb}' (\textit{`book'}) $+$ `\textit{cog}' (\textit{`table'}), usually orthographically marked with a hyphen and morphologically marked with the ezafe proclitic on the compound head or rightmost item.
    \item \textbf{Opaque Loans.} To simulate superstratal influence~\citep{kahane1986typology}, a controlled portion of items is adapted from archaic or less familiar sources, e.g., reconstructed forms, whose phonological shapes differ from the native stock. 
    These items model historically plausible borrowing while remaining difficult for models to identify by external knowledge trivially.
    \item \textbf{Transparent loans.} 
    To reflect cross-linguistic diffusion of modern concepts~\citep{wierzbicka2006concept}, a small set of items is adapted from major languages, e.g., \textit{rajdiw} `radio', remaining recognisable in meaning despite phonological adaptation. 
    Together with opaque loans, this yields a realistic loan profile.
   
\end{itemize}  

\paragraph{Semantic Mapping Policy.}
We do not engineer semantics to be adversarial. 
Instead, where possible, {Camlang} lexemes map one-to-one to English equivalents. 
Nevertheless, motivated sense splits or mergers are permitted, e.g., \textit{ṇat} `play (with toys)' vs.\ \textit{get} `play (sports/instruments)' and \textit{ǵomluṇ} (`usually/generally'), reflecting attested partitions of semantic space via metaphor/metonymy.

\paragraph{Dictionary Structure and Readability.}
Entries are alphabetised by the \emph{underlying form} (the morphophonological base). 
For each entry, we record: 
(1) English gloss (and relevant subcategorisation, e.g., genitive vs.\ accusative complements); 
(2) part of speech and honorific status; 
(3) underlying form; 
(4) surface citation form; and 
(5) etymology. 
(5) is not released to models to avoid leakage of external knowledge. (3) and (4) anchor morphological analysis by distinguishing underlying from surface realisation.

Since {Camlang} is morphologically rich, many surface forms are not listed verbatim.
Thus, the dictionary's underlying forms serve as lemmata for analysis and retrieval, both for humans and models.

\subsubsection{Iterative Refinement}\label{sec:iterative_construction}
Starting from the initial grammar book and dictionary, we perform iterative refinements with three aims:
(1) to improve learnability and typological plausibility by removing unnecessary complexities and outliers;
(2) to expand the lexicon and specify constructions required for naturalistic expression of complex meanings; and
(3) to simulate {Camlang}'s plausible historical development, so that new features arise via naturalistic grammaticalisation rather than arbitrary design.

In each round, four annotators (linguistics undergraduates from Cambridge and Oxford who have experience in constructed languages and L2 acquisition) independently translated a set of English sentences or dialogues using the current grammar and dictionary. 
The source material increases in form and expressive complexity across rounds: starting with simple declaratives (e.g., ROCStories~\citep{mostafazadeh-etal-2016-corpus}), moving to short daily dialogues, and culminating in longer interview-style exchanges.\footnote{\url{https://www.ox.ac.uk/sites/files/oxford/media_wysiwyg/Interview\%20Podcast\%20-\%20Transcript.pdf}}
Translation divergences are then used diagnostically to locate missing constructions, underspecified rules, or structural inconsistencies.
The discussion also leads to fixes incorporated into the next grammar and dictionary revision. 
Note that differences across translations are not treated as errors, since {Camlang} permits paraphrastic variation and multiple acceptable syntactic and pragmatic realisations. 
Only divergences indicating under-specification or inconsistency trigger revisions. 
For lexical expansion, new stems are added only when required to express concepts absent from the current dictionary. 
After refinement, we further manually review and provide feedback on contradictions or underspecified interactions.
We stop after Round~3 when no further feedback beyond minor editorial clarifications is provided.
Across all rounds, each annotator works with a total of 2,263 tokens, 737 types, and 207 sentences.
While modest in size, this corpus fits our purpose to assess whether the grammar–dictionary specification affords adequate coverage of naturalistic expressions across diverse syntactic structures and discourse contexts. 

Here, we summarise the major refinements to the grammar book in each round:
\begin{itemize}
    \item\textbf{Round 1.} 
    The discussion and refinement focus on surface-form consolidation and core nominal/clausal regularisation. 
    The refinement mainly addresses the standardisation of orthography–phonology conventions~\citep{venezky2013role} and re-organisation of the nominal system around simple exponence, together with the grammaticalisation of oblique pronouns into pronominal clitics~\citep{borer2020syntax} (remaining disharmonic) and a reanalysis of the accusative/genitive contrast with postpositions as a dynamic–stative opposition~\citep{koopman2010prepositions}.
    We also fix the basic case assignment and ordering for double-complement predicates.

\item\textbf{Round 2.} 
This round focuses on systematising the behaviour of clitics, reworking some high-frequency morphemes' forms while adding sources of stylistic variation.
In particular, we specify clitic stacking and movement rules explicitly~\citep{roberts2010agreement}, and streamline the core paradigms (pronouns, TAM, information-structure and interrogative markers) to improve distinctiveness and clarity.
In addition, we adopt limited periphrastic resources (\textit{do}-support and light verbs), and replace originally arbitrary evidential contrast and agreement morphology with diachronically motivated ones, while introducing a vowel–glide dissimilation rule to increase phonological opacity~\citep{mccarthy1999sympathy}. 
When annotating the dialogues, we recognise that the lack of a pragmatic architecture limited the stylistic scope of {Camlang}. 
In response, we devise a topic-based honourific register~\citep{agha2006language}, introducing a level of controlled pragmatic application to account for the range of contexts in which a conversation might take place~\citep{parsons2017systematic}.

\item\textbf{Round 3.} 
We mainly demarcate functions of some grammatical categories and advanced {Camlang}'s macro-typology in a diachronically naturalistic way~\citep{greenberg1995diachronic}.
In particular, we separate the definiteness function from the linker by introducing an independent definite marker, and introduce full reduplication to meet typological expectations~\citep{haspelmath2005world}.
We further establish a first-person inclusive/exclusive split and make specificity-driven Differential Object Marking explicit~\citep{rodriguez2007syntax}. 
We codify situations of use of mood, modality, evidentiality and epistemic markers.
And we turn {Camlang} from agglutinative to polysynthetic via grammaticalisation of pronominal clitics into obligatory head-markers.\footnote{A historically canonical development captured as a morphosyntactic cycle \citep{+1905+285+313}.}
We also structured mathematical expressions for future applications. 

\end{itemize}

After refinement, we also add a \textit{General Description} section, which summarises the language's major features.
The resulting grammar book contains 453 sentences and 9,477 tokens.

For the dictionary, new entries are added on demand for texts, and forms are amended only when grammar changes (e.g., orthography or derivational rules) impact citation forms. 
At the time of experiments, the dictionary contains 1{,}511 entries.
We show the distribution of the lexical sourcing and formation as in Table~\ref{tab:sourcing_distribution}.

\begin{table}[t]
    \centering
    \begin{tabular}{l|cc}
    \toprule
        \textbf{Category} & \textbf{\#}  & \textbf{\%}\\
        \midrule
        Native Word & 698 & 46.19\\
        Derived Word & 588 & 38.91 \\
        Compound Word & 103 & \ \ 6.82 \\
        Opaque Loan & \ \ 80 & \ \ 5.29\\
        Transparent Loan & \ \ 42&\ \ 2.78\\
        \bottomrule
    \end{tabular}
    \caption{The distribution of lexical sourcing and formation as in the dictionary by the experiment time.}
    \label{tab:sourcing_distribution}
\end{table}

Using the annotated data from the above procedure, we also measure the cross-annotator consistency across the three rounds as an indirect indicator of {Camlang}'s learnability under an explicit grammar and dictionary (metrics and results in \S~\ref{sec:camlang_analysis}).


\subsection{The Overview of {Camlang}}\label{sec:overview_camlang}

\textit{To avoid potential data leakage to LLMs, we do not release the full grammar book and dictionary at this stage. Instead, we provide a high-level overview of the linguistic profile of \textit{{Camlang}}. We also provide a profile of the World Atlas of Language Structures (WALS)-style features in Appendix~\ref{sec:wals}.}

\paragraph{Phonology.}
{Camlang} has 50 consonantal phonemes, among which 30 do not appear in native citation forms, but result from loaning or consonant mutation, an important morphophonological process that operates at cliticisation boundaries. The inventory is largely symmetric with four places of articulation, and each phoneme can appear at both onset and coda. {Camlang} has a Turkic-type harmonic 8-vowel system with vowel reduction, which neutralises contrast between three vowels at unstressed positions. {Camlang} has strict (C)V(C) phonotactics that prohibits adjacent vowels, often satisfied by epenthesis.
The combination of these features means learners of {Camlang} must apply phonological rules in the specified sequence in order to derive surface forms from underlying forms or reversely. For example:
\ex\label{phonexample}
\begingl
\glpreamble \textit{lichéwcymyśür}//
\gla Segmented: li- chew {} -cy -my -śür {}//
\glb Morphemic: lI= x= cew -RED $-mA^4$ -s =jUr//
\glb Gloss: 2SG= EZ= answer -PROG -NMLS -GEN =at//
\glft `when you are answering'//
\endgl
\xe
The {Camlang} word in Case \ref{phonexample} consists of seven morphemes, none of which appear in underlying form. 
Learners cannot comprehend or produce this word if they do not apply relevant phonological rules, because the ungrammatical result of simple morphemic concatenation, \textit{*lixcéwcemasjur}, would never appear.

\paragraph{Morphosyntax.}
{Camlang} is a polysynthetic language with obligatory head-marking morphology, consistent pro-drop, and predominant simple exponence. 
For derivation and inflection, {Camlang} uses both affixes and clitics extensively. {Camlang} has a slight preference for suffixation over prefixation and circumfixation, and a slight preference for procliticsation over encliticisation. 
{Camlang}'s nouns are marked for many grammatical functions, such as definiteness, plurality, and case, and can take postpositional enclitics. 
Their possessors are marked with proclitics. {Camlang}'s verbs are marked categories such as for valency, aspect, tense, evidentiality, mood, and finiteness. 
Finite subjects are marked on verbs with agreement suffixes while other arguments are marked with proclitics. 
Nonfinite verb forms like nominalisations are extensively used in complex constructions realised as subordination in English.

The combination of these features means learners of {Camlang} must use morphosyntactic knowledge to analyse syntactic dependencies and grammatical functions when comprehending or producing agglutinative word forms, which is more challenging for learners with analytic mother tongue languages or training languages like English and Mandarin Chinese. 
For example, the written word in Case~\ref{phonexample} contains semantic information equivalent to that of an English subordinate clause due to its morphological complexity, and would be incomprehensible if the learner fails to perform morphological segmentation and analysis according to specific grammar points.

Due to the heavy functional load taken by {Camlang}'s morphology, its syntactic configurations are relatively simple. 
Structurally, all projections lower than the Complementiser Phrase \footnote{for Complementiser Phrase, see Chapter 9 of \citet{radford2004english}} in {Camlang} are head-final, giving SOV default word order. 
The active left-periphery hosts the information-structural functions of topicalisation and focalisation, and other word orders are used in many marked clause types. 
Otherwise, word orders are usually fixed and decided by the hierarchical structure. 

\paragraph{Sociopragmatics.} 
{Camlang} does not exhibit a T/V distinction in second-person reference forms~\citep{8407211f-5665-3fa0-bfd2-5965615377de}, but showcases lexicalised register as found in languages like Thai ~\citep{alma990158258210107026}, with some concepts being denotable by an ordinary linguistic variant and an honorific linguistic variant.
Moreover, morphosyntactic constructions may be used as stylistic devices to give the text a more native, natural style as opposed to a more foreign, translationese style. 
{Camlang}'s sociopragmatic description is currently underdeveloped, but we hope to enrich it from this basis and use it to explore discourse-related tasks.


\subsection{Analysis and Evaluation on {Camlang}}
\label{sec:camlang_analysis}

\subsubsection{Grammar Clarity and Practical Learnability}
We assume that if the grammar--dictionary specification is sufficiently explicit and internally consistent, then the {Camlang} learners will converge on the same morphological choices and on similar clause-level linearization for a given English input~\citep{reber1967implicit,ELLIS2005209,lake2018generalization}. 
In \S~\ref{sec:iterative_construction}, we iteratively refine the grammar and dictionary through three rounds of translation by four annotators.
We thus analyse the cross-annotator consistency as \emph{indirect} evidence about the language learnability.\footnote{Our use of ``\textit{learnability}'' is empirical, different from classical notions, such as \citep{gold1967language} and \citep{valiant1984theory}. In L2 acquisition, the learnability/teachability hypothesis has also been discussed \citep{pienemann1989language}. We do not claim formal identifiability here and focus on practical learnability under explicit instruction.}
We calculate the averaged \textsc{Rouge}~\citep{lin2004rouge} scores between different annotators at different rounds, and report the results at two granularities.
At the word-level, we tokenise the {Camlang} sentences by spaces.
At the morpheme-(surface morph) level, we manually tokenise words into morphemes according to the grammar/dictionary. 
The results are shown in Table~\ref{tab:rouge_by_round_multiheader}.

\begin{table}[t]
\centering
\setlength{\tabcolsep}{6pt}
\begin{tabular}{c|ccc|ccc}
\toprule
\multirow{2}{*}{\textbf{Round}} & \multicolumn{3}{c|}{\textbf{Word}} & \multicolumn{3}{c}{\textbf{Morpheme}} \\
\cmidrule(lr){2-4} \cmidrule(l){5-7}
 & \textbf{R1} & \textbf{R2} & \textbf{RL} & \textbf{R1} & \textbf{R2} & \textbf{RL} \\
\midrule
1 & 46.23 & 19.09 & 44.27 & 67.51 & 36.27 & 61.89 \\
2 & 57.96 & 33.69 & 56.67 & 72.97 & 50.35 & 68.39 \\
3 & 62.91 & 40.99 & 60.10 & 75.04 & 55.40 & 69.88 \\
\bottomrule
\end{tabular}
\caption{Averaged ROUGE scores across annotators by round at word and morpheme levels.}
\label{tab:rouge_by_round_multiheader}
\end{table}

Generally, from Round~1 to Round~3, \textsc{Rouge} increases consistently at both granularities (word: R1 46.23$\to$62.91, R2 19.09$\to$40.99, RL 44.27$\to$60.10; morpheme: R1 67.51$\to$75.04, R2 36.27$\to$55.40, RL 61.89$\to$69.88).
The largest gains in \textsc{Rouge}-2 indicate increasing stability of \emph{local} combinations (affix order, clitic/particle placement), while gains in \textsc{Rouge}-L point to a more constrained \emph{global} clause linearisation.
These trends persist despite rising source complexity across different rounds (from simple declaratives to longer complex dialogues), suggesting that the grammar–dictionary specification became operationally clearer and thus affords practical learnability under explicit instruction.
Morpheme-level scores are consistently higher than word-level ones (e.g., Round~3 RL 69.88 vs.\ 60.10).
Decomposing words exposes shared functional structure (affixes, clitics, and morphotactic slots) and reduces lexical sparsity, thereby increasing the chance of $n$-gram overlap across independently produced translations. 
This is in line with our morphologically rich design.
Accordingly, higher morpheme-level \textsc{Rouge} indicates convergence on the same grammatical machinery (choice of functional morphemes and slot order) rather than mere coincidence at the word level. 
It should be noted that, because EM metrics are boundary-convention–sensitive, morpheme-level \textsc{Rouge} should be read as a conservative estimate.

We acknowledge that the present analysis only considers the mapping from English into {Camlang}, i.e., the production aspect. 
A fuller account of practical learnability would also require examining comprehension in the opposite direction, i.e., the comprehension aspect~\citep{vanpatten2004processing,pickering2013integrated}.
We complement it later in \S~\ref{sec:experiments}, where one human participant is asked to solve the commonsense question answering task given the grammar book and dictionary, and achieves an accuracy around 87\%, thereby addressing this aspect of bidirectional learnability.

\subsubsection{Novelty and Typological Positioning}

\begin{table}[t]
\centering
\small
\setlength{\tabcolsep}{6pt}
\begin{tabular}{l c c c c c c}
\toprule
\textbf{Rank} & \textbf{Code} & \textbf{Similarity} & \textbf{$|F_X \cap F_Y|$} & \textbf{Name} & \textbf{Genus} & \textbf{Family}  \\
\midrule
\multicolumn{7}{c}{\textbf{$\geq$20 overlap}}\\
\midrule
Top--1 & awa & 0.96 & 27 & Awadhi & Indic & Indo-European\\
Top--2 & gam & 0.95 & 20 & Gamo & Ta-Ne-Omotic & Afro-Asiatic \\
Top--3 & kum & 0.91 & 23 & Kumauni & Indic &  Indo-European \\
Top--4 & knn & 0.89 & 28 & Kinnauri & Bodic & Sino-Tibetan \\
Top--5 & bho & 0.82 & 33 & Bhojpuri & Indic & Indo-European \\
\midrule
\multicolumn{7}{c}{\textbf{$\geq$30 overlap}}\\
\midrule
Top--1 & bho & 0.82 & 33 & Bhojpuri & Indic & Indo-European \\
Top--2 & sid & 0.81 & 32 & Sidaama & H. East Cushitic	& Afro-Asiatic\\
Top--3 & sna & 0.80 & 30 & Shina &	Indic & Indo-European\\
Top--4 & guj & 0.75 & 36 & Gujarati	& Indic &	Indo-European\\
Top--5 & pba & 0.75 & 32 & Pima Bajo & Tepiman	& Uto-Aztecan\\
\midrule
\multicolumn{7}{c}{\textbf{$\geq$40 overlap}}\\
\midrule
Top--1 & dji & 0.74 & 42 & Djingili	&	Djingili &	Mirndi\\
Top--2 & raw & 0.70 & 40 & Rawang &	Nungish	 & Sino-Tibetan \\
Top--3 & tvo & 0.68 & 41 & Tatar &	Turkic	& Turkic\\
Top--4 & mns & 0.68 & 59 & Mansi &	Ugric	& Uralic \\
Top--5 & bel & 0.67 & 43 & Belhare &	Himalayish &	Sino-Tibetan\\
\midrule
\multicolumn{7}{c}{\textbf{$\geq$50 overlap}}\\
\midrule
Top--1 & mns & 0.68 & 59 & Mansi &	Ugric	& Uralic \\
Top--2 & kty & 0.67 & 66 & Khanty &	Ugric	& Uralic\\
Top--3 & ttu & 0.67 & 51 & Tsova-Tush	&	Nakh &	Nakh-Daghestanian\\
Top--4 & pan & 0.66 & 65 & Panjabi	&	Indic &	Indo-European\\
Top--5 & dgr & 0.65 & 55 & Dagur &	Mongolic & Mongolic\\
\bottomrule
\end{tabular}
\caption{Camlang's top neighbours with $\geq$20, $\geq$30, $\geq$40, and $\geq$50  overlapping features. 
High similarities at this threshold are based on very small overlaps. Results on $\geq$40 and $\geq$50 features are more stable and typologically meaningful. H. East Cushitic is Highland East Cushitic.}
\label{tab:camlang_comparison}
\end{table}

Following~\citep{georgi-etal-2010-comparing}, we adapt a WALS-style feature-based method to compare the similarity between two languages. 
WALS~\citep{haspelmath2005world} is the most widely used and largest typological database, covering 2,662 languages with 144 chapters of features across domains such as phonology, morphology, and word order.\footnote{\url{https://wals.info/chapter/s1}}
Since typological inventories are sparse and some features are not directly comparable across languages, we restrict comparison to the intersection of features available for both languages. 
We manually construct a WALS-style typological profile for {Camlang} based on its grammar and dictionary, aligning each relevant structural property with the corresponding WALS feature value (Appendix~\ref{sec:wals}).
Overall, Camlang contains 134 features.

\begin{table}[t]
\centering
\small
\setlength{\tabcolsep}{6pt}
\begin{tabular}{l c c c c c c}
\toprule
\textbf{Rank} & \textbf{Code} & \textbf{Similarity} & \textbf{$|F_X \cap F_Y|$} & \textbf{Name} & \textbf{Genus} & \textbf{Family}  \\
\midrule
\multicolumn{7}{c}{\textbf{Awadhi's (Indic, Indo-European) top neighbours $\geq$20 overlap}}\\
\midrule
Top--1 & ttu & 1.00 & 27 & Tsova-Tush &	Nakh& Nakh-Daghestanian \\
Top--2 & knn & 1.00 & 25 & Kinnauri & Bodic	& Sino-Tibetan \\
Top--3 & sna & 1.00 & 25 & Shina	&	Indic	& Indo-European \\
Top--4 & guj & 1.00 & 21 & Gujarati	&	Indic	& Indo-European \\
Top--5 & kty & 0.96 & 30 & Khanty	&	Ugric	& Uralic \\
\multicolumn{7}{c}{...}\\
Top--8 & - & 0.96 & 27 & Camlang & - & - \\
\midrule
\multicolumn{7}{c}{\textbf{Bhojpuri's (Indic, Indo-European) top neighbours $\geq$30 overlap}}\\
\midrule
Top--1 & hin & 0.97 & 33 & Hindi& Indic	& Indo-European \\
Top--2 & hop & 0.91 & 33 & Hopi	& N. Uto-Aztecan &	Uto-Aztecan \\
Top--3 & kor & 0.90 & 30 & Korean & Korean &	Korean \\
Top--4 & ttu & 0.90 & 30 & Tsova-Tush &	Nakh & Nakh-Daghestanian\\
Top--5 & mgg & 0.84 & 31 & Mangghuer &	Mongolic & Mongolic \\
\multicolumn{7}{c}{...}\\
Top--9 & - & 0.82 & 33 & Camlang & - & - \\
\midrule
\multicolumn{7}{c}{\textbf{Djingili's (Djingili, Mirndi) top neighbours $\geq$40 overlap}}\\
\midrule
Top--1 & kor & 0.74 & 46 & Korean & Korean &	Korean  \\
Top--2 & - & 0.74 & 42 & Camlang & - & - \\
Top--3 & hin & 0.71 & 48 & Hindi & 	Indic	& Indo-European\\
Top--4 & ket & 0.71 & 41 & Ket & 	Yeniseian & 	Yeniseian \\
Top--5 & bsq & 0.70 & 47 & Basque & Basque & Basque \\
\midrule
\multicolumn{7}{c}{\textbf{Mansi (Urgic, Uralic) top neighbours $\geq$50 overlap}}\\
\midrule
Top--1 & ntu & 0.77 & 51 & Nenets & Samoyedic & Uralic  \\
Top--2 & qim & 0.74 & 60 & Quechua 	&	Quechuan  &	Quechuan \\
Top--3 & geo & 0.72 & 58 & Georgian	 &	Kartvelian&	Kartvelian\\
Top--4 & ket & 0.71 & 52 & Ket & 	Yeniseian & 	Yeniseian \\
Top--5 & ain & 0.69 & 55 & Ainu	& Ainu	& Ainu \\
Top--6 & - & 0.68 & 59 & Camlang & - & - \\
\bottomrule
\end{tabular}
\caption{Top neighbours of Camlang compared with their own closest languages under the same overlap thresholds. N. Uto-Aztecan is Northern Uto-Aztecan. }
\label{tab:neighbours_comparison}
\end{table}

Let $F_X$ and $F_Y$ be the sets of features encoded for languages $X$ and $Y$, respectively. 
We restrict comparison to their overlap $F_X \cap F_Y$, in order to avoid spurious differences due to missing values and to ensure
comparability.
For each feature $f$ in the overlap, we assign an indicator score:
\begin{equation}
\delta_f(X,Y) =
\begin{cases}
1, & \text{if } v_f(X) = v_f(Y), \\
0, & \text{otherwise},
\end{cases}
\end{equation}
where $v_f(X)$ and $v_f(Y)$ are the values of feature $f$ in languages $X$ and $Y$, respectively. 
The overall similarity is then defined as the proportion of matched features:
\begin{equation}
\text{sim}(X,Y) = \frac{\sum_{f \in F_X \cap F_Y} \delta_f(X,Y)}{|F_X \cap F_Y|}.
\end{equation}
Following~\citet{georgi-etal-2010-comparing}, we set a minimum overlap threshold and report results across different settings (20, 30, 40 and 50 features).
For comparison, we compute the similarities between each top neighbour of Camlang and all other languages under the same minimum-overlap setting, and the similarities between Camlang and high-resource languages.

Table~\ref{tab:camlang_comparison} presents Camlang’s closest neighbours under different feature-overlap thresholds. 
At lower thresholds ($\geq$20, 30), Camlang appears close to Indic languages such as Awadhi, Kumauni, and Bhojpuri, as well as Afro-Asiatic and Sino-Tibetan languages. 
However, these results are based on small overlaps and are less stable. 
More meaningful comparisons emerge at higher thresholds ($\geq$40, 50), where Camlang’s closest neighbours include Djingili (Mirndi), Rawang (Sino-Tibetan), Tatar (Turkic), Mansi (Ugric), and Dagur (Mongolic), with similarities in the 0.67--0.74 range. 
These results suggest that Camlang does not collapse into any single language family, but rather occupies a mixed region of typological space, combining features from multiple families in a stable and coherent way.

Table~\ref{tab:neighbours_comparison} further compares Camlang’s top-1 neighbours against their own closest languages. 
Natural languages such as Awadhi (an Indo-Aryan language) and Bhojpuri (also an Indo-Aryan language) have higher similarities with their genealogical relatives (e.g., Indic languages), with near-perfect similarity scores. 
Conversely, in cases such as Djingili and Mansi, Camlang itself ranks among their top neighbours, often alongside typologically distant families such as Korean, Basque, or Quechuan. 
This contrast highlights that Camlang is sufficiently naturalistic to be typologically comparable to real languages, yet it is distinct in that it spans across multiple genealogical groups rather than grouping tightly with one family.

\begin{table}[t]
\centering
\small
\setlength{\tabcolsep}{6pt}
\begin{tabular}{l c c c c c}
\toprule
\textbf{Code} & \textbf{Similarity} & \textbf{$|F_X \cap F_Y|$} & \textbf{Name} & \textbf{Genus} & \textbf{Family}  \\
\midrule
eng & 0.44 & 131 & English &	Germanic & Indo-European \\
fre & 0.42 & 130 & French & Romance & Indo-European \\
tur & 0.63 & 133 & Turkish & 	Turkic & 	Turkic \\
mnd & 0.45 & 128 & Mandarin & Chinese & Sino-Tibetan \\
jpn & 0.49 & 126 & Japanese &	Japanese& Japanese \\
\bottomrule
\end{tabular}
\caption{Comparison between Camlang and high-resource languages. }
\label{tab:high_resource_comparison}
\end{table}

Table~\ref{tab:high_resource_comparison} shows that Camlang maintains relatively low similarity with major high-resource languages such as English (0.44), French (0.42), Mandarin (0.45), and Japanese (0.49), confirming that it does not trivially overlap with the typological profiles of widely studied languages. 
Camlang shows a higher similarity with Turkish (0.63, based on 133 overlapping features). 
This is consistent with our design that Camlang incorporates typological traits such as vowel harmony and agglutinative morphology that are characteristic of Turkic languages (\S~\ref{sec:grammar_book}). 
Taken together, these results demonstrate that while Camlang remains distinct from common high-resource languages, it naturally aligns more closely with Turkic languages, reinforcing both its novelty as a constructed language.

\subsubsection{Richness Analysis}\label{sec:richness}

\begin{figure}
    \centering
    \includegraphics[width=1\linewidth]{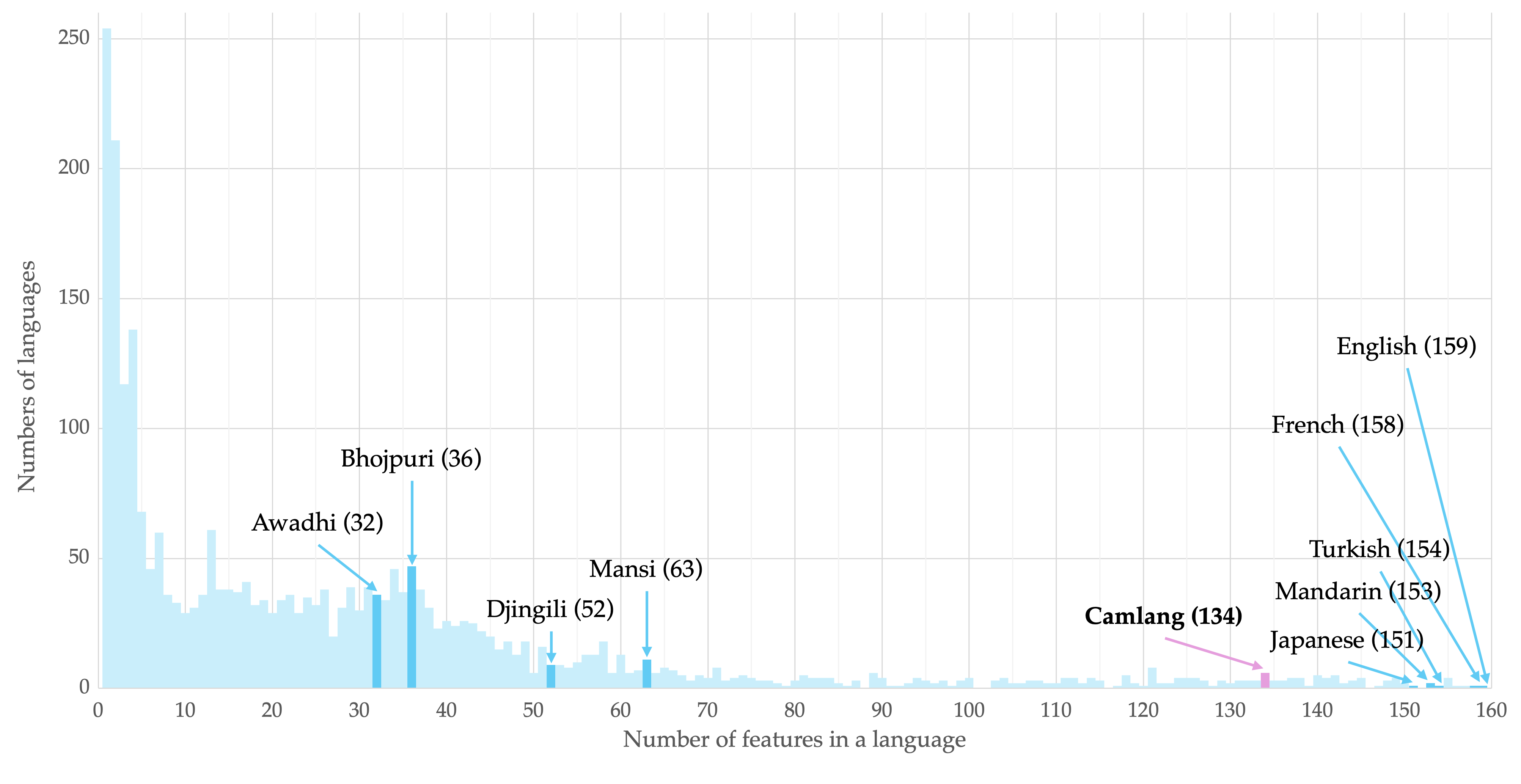}
    \caption{Distribution of WALS features in documented languages (with selected languages highlighted).}
    \label{fig:feature_distribution}
\end{figure}

We measure the \textit{richness} of Camlang by comparing its WALS features with other languages. 
Figure~\ref{fig:feature_distribution} shows the distribution of feature counts across documented languages. 
We see that most languages are annotated with only a few dozen features, which can be attributed to limited documentation and scarce resources for many low-resource and endangered languages. 
As a result, their typological richness in WALS is insufficiently captured. 
Using such languages for evaluation would cannot provide a comprehensive test of linguistic competence and risk biasing the evaluation towards a narrow subset of phenomena.
In contrast, Camlang has 134 features, placing it above 97.52\% of WALS languages.
Compared to high-resource languages such as English (159), French (158), and Mandarin (153), while Camlang does not aim to replicate the full complexity of natural languages, it ensures sufficient typological richness to support systematic evaluation.


\section{The {Camlang-CSQA-v0} Test Set}

With the grammar book and dictionary stabilised, we construct {Camlang-CSQA-v0}, a small-scale multiple-choice dataset of commonsense questions in the {Camlang} language, derived from {CommonsenseQA}~\citep{talmor2019commonsenseqa}. 

We choose CommonsenseQA~\citep{talmor2019commonsenseqa} as the source task because it is a well-established benchmark for evaluating commonsense reasoning in NLP, consisting of daily scenarios that require integrating world knowledge and simple reasoning. 
Moreover, prior work has shown that the knowledge required to answer CommonsenseQA is already well represented in current LLMs~\citep{aggarwal2021explanations,krause2023commonsense,bian2024chatgpt}.  
By translating the task into {Camlang}, the challenge no longer lies in retrieving commonsense knowledge itself, but in whether models can correctly interpret the grammar book and dictionary and apply explicit rules to integrate this unfamiliar linguistic input with their existing commonsense knowledge.

\subsection{Task Definition}

The task is multiple-choice question answering in {Camlang}.  
Each instance consists of a commonsense question and six candidate answers, including one ``\textit{none of the above}'' option.
The systems are required to select the single correct answer based on the {Camlang} input, with access to the grammar book and dictionary.


\subsection{Data Selection and Pre-processing}

We first randomly select 50 instances from the CommonsenseQA test set.\footnote{\url{https://huggingface.co/datasets/tau/commonsense_qa/}}  
In practice, we observe that some instances are either less reasonable or contain ambiguous answer options. 
We first modify or replace such options to ensure that the correct answer is more reasonable. 
Instances with ambiguous questions are removed.
Another problem is that the {Camlang} dictionary is limited and cannot cover all entities in CommonsenseQA.
To avoid continuously expanding the dictionary with low-frequency words, we thus replace or modify such entities with semantically similar ones already present in the dictionary, if they are not essential for reasoning or answer selection.  
For example, given the question ``\textit{How could a weasel go to another continent?}'' with the correct answer ``\textit{by ship}'', we replace ``\textit{weasel}'' with ``\textit{cat}'', which is already covered by the dictionary.  
In addition, we insert an extra option ``\textit{none of the above}'' into each instance. 
After this adjustment, we manually select the gold answers, since the CommonsenseQA test set does not provide them. 
The resulting set of multiple-choice instances is then translated from English into {Camlang}, yielding a cleaner set of 47 questions.

\subsection{Annotation}

Given the English questions and multiple-choice options, we ask the core annotator, who participates in Camlang development, to translate them into {Camlang}, with reference to the grammar book and dictionary. 
The translation is not required to follow English syntax in a word-to-word manner; instead, the annotator is allowed to adapt the sentences according to the typological characteristics of {Camlang}. 
When no reasonable substitution is available in the existing dictionary, the annotator extends the lexicon with new entries.  
In addition to the translated sentences, we also provide gloss-level annotation, which specifies the morpheme-by-morpheme alignment and thereby ensures both accuracy and interpretability of the annotation.  
For example, given the question ``\textit{How could a cat go to another continent?}'', we interpret the gloss as ``\textit{cat=TOP how go-INF=can another continent-GEN?}''.
After the initial annotation, a second annotator, who is also involved throughout the grammar design process, reviews and revises the translations to ensure consistency and correctness.

\subsection{Dataset Statistics}

\begin{figure}[t]
    \centering
    \includegraphics[width=0.73\linewidth]{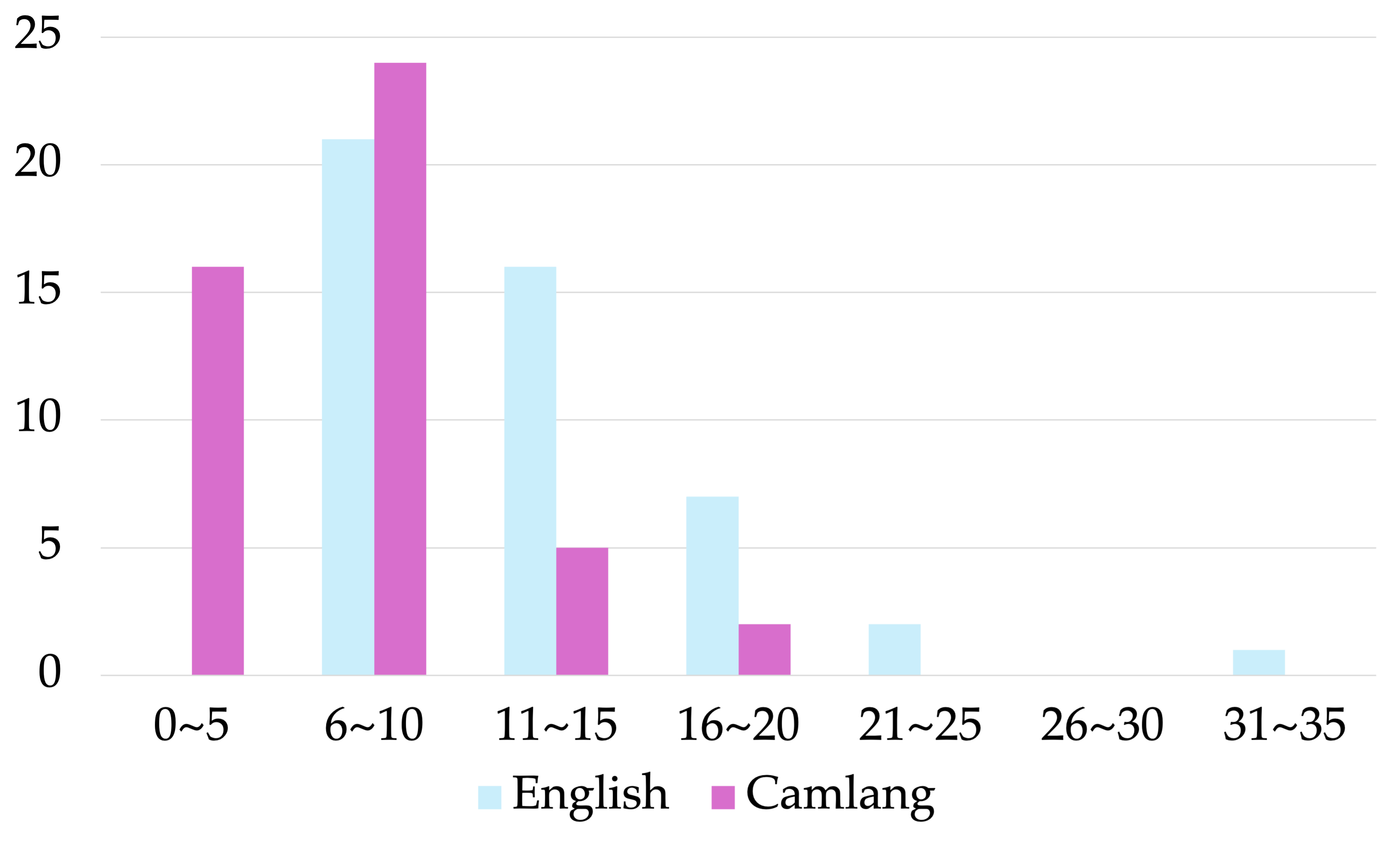}
    \caption{Word count distribution of questions in English and {Camlang}. The $x$-axis indicates the number of words per question, and the $y$-axis indicates the number of questions.}
    \label{fig:length_distribution}
\end{figure}

We analyse the length of questions before and after translation. 
Lengths are measured in words, where both English and {Camlang} are tokenised by spaces.
The distribution is shown in Figure~\ref{fig:length_distribution}.
On average, English questions contain 12.34 words, whereas their {Camlang} counterparts are shorter with 7.32 words.  

We observe three main trends.  
First, both English and {Camlang} questions are concentrated between 6–15 words, reflecting the original distribution of CommonsenseQA.  
Second, {Camlang} exhibits a higher proportion of very short questions (0–5 words), owing to its morphological compactness: information that requires separate function words in English (e.g., auxiliaries, prepositions) is often expressed via affixes or clitics in {Camlang}.  
Third, English questions extend to much longer tails (over 20 words), while no {Camlang} question exceeds 20 words.  
This indicates that translation into {Camlang} systematically yields more compact formulations, while preserving the semantic content needed for reasoning.



\section{Experiments}\label{sec:experiments}
With the grammar book, dictionary, and {Camlang-CSQA-v0}, we conduct experiments to evaluate both human and model performance.  
This pilot experiment serves two main purposes.  
First, we assess the {Camlang} evaluation paradigm by testing whether human participants, given only the grammar book and dictionary, can comprehend questions encoded in {Camlang}, thereby validating that our method supports metalinguistic deductive learning.  
Second, we investigate whether LLMs can interpret the explicit grammatical rules and apply them to perform deductive reasoning in an unfamiliar language.

\subsection{Experimental Setup}
We conduct experiments on {Camlang-CSQA-v0} in a zero-shot setting.

\subsubsection{Baseline Systems and Implementation Details}

We consider strong reasoning models as our baseline systems. 
For comparison, we also include a general-purpose model and human participant:
\begin{itemize}
    \item \textbf{GPT-4o} (\texttt{2024-11-20}): A general-purpose LLM from OpenAI~\citep{openai2024gpt4o}. 
    \item \textbf{GPT-o4-mini} (\texttt{2025-04-16}): A lightweight, efficient reasoning LLM~\citep{openai2024gpt3o}.
    \item \textbf{GPT-o3} (\texttt{2025-04-16}): OpenAI's most capable model in the o-series of reasoning models~\citep{openai2024gpt3o}.
    \item \textbf{GPT-5} (\texttt{2025-08-07}): The most recent LLMs from OpenAI, featuring substantial improvements in coding and reasoning~\citep{openai2025gpt5}. 
    \item \textbf{DeepSeek-R1} (\texttt{2025-05-28}): An open-source reasoning LLM from DeepSeek, focusing on strong logical and reasoning capabilities~\citep{deepseekai2025deepseekr1}.
    \item \textbf{Human Participant}: A first-year undergraduate student in Linguistics at Oxford.
\end{itemize}

For OpenAI models, we consider two configurations enabled by the API interface: (1) Tools-enabled: we upload the grammar book (\texttt{.md}) and the dictionary (\texttt{.xlsx}) as files and allow the model to call \texttt{File Search} and \texttt{Code Interpreter}, respectively, and;
(2) Context-only: we prepend the full text of the grammar book and a textual export of the dictionary to the input as context, with tools disabled. 
For {DeepSeek-R1}, we use the context-only configuration as it does not support such tools.

For reasoning models ({o3}, {o4-mini}, and {GPT-5}), we set the \texttt{reasoning effort} to \texttt{high}. 
For \textit{GPT-4o}, we set \texttt{temperature} and \texttt{top-p} to 0. 
For the other models, these two parameters are not exposed, and we therefore use the default configuration.
We add prompts before the question to instruct the LLMs to explicitly make use of the grammar book and dictionary, and conclude its output with a sentence of the form \textit{``The final answer is X''}, where $X$ is an option (e.g., \textit{``A''} or ``\textit{B}''). 
The detailed prompts are shown in Appendix~\ref{sec:prompts}.
In addition to choosing an answer option, the human participant is asked to write explanations to demonstrate their choice, simulating the reasoning steps of LLM, which is later used for evaluation and analysis.

We also run the systems on the English version of our testing instances without the grammar book and dictionary to provide a controlled comparison.
For human participant, to avoid leakage, we ask the annotator to complete the English questions after the Camlang questions.

\subsection{Exact Match Results}

\subsubsection{Evaluation Method}
Following standard practice in reasoning benchmarks, we report average accuracy using exact matching (EM).
In this setting, a prediction is considered as correct if and only if the model's output contains a final answer of the form 
\textit{``The final answer is X''}, and the value $X$ exactly matches the gold option. 
If a response contains multiple such phrases, we take the last one as the final prediction.

\subsubsection{Result Discussion}

We first discuss the EM results as shown in Table~\ref{tab:csqa-result}.
All observations in this subsection are based on the EM accuracy only, which should not be interpreted as direct evidence of metalinguistic deductive reasoning.

\begin{table}[t]
\centering
\setlength{\tabcolsep}{8pt}
\begin{tabular}{l cc cc c}
\toprule
\multirow{2}{*}{\textbf{Model}} & \multicolumn{2}{c}{\textbf{Camlang}} & \multicolumn{2}{c}{\textbf{English}} & \multirow{2}{*}{\textbf{ACC $\downarrow$}} \\
\cmidrule(lr){2-3} \cmidrule(lr){4-5}
 & \textbf{ACC} (\%) & \textbf{Time} (s) & \textbf{ACC} (\%) & \textbf{Time} (s) &  \\
\midrule
4o (tool)     & 21.28 &\ \ 75.04 & \multirow{2}{*}{89.36} & \multirow{2}{*}{19.49} & 68.08 \\
4o (context)  & 21.28 &\ \ 31.98 &  &  & 68.08 \\
\midrule
DS (context)  & 40.43 & 263.63   & 85.11 &  58.70 & 44.68 \\
\midrule
o4-mini (tool)& 25.53 & 413.84   & {91.49} & {14.20} & 65.96 \\
\midrule
o3 (tool)     & 27.66 & 566.09   & \multirow{2}{*}{97.87} & \multirow{2}{*}{\ \ 7.86} & 70.21 \\
o3 (context)  & 46.81 & 158.23   &  &  & 51.06 \\
\midrule
GPT-5 (tool)   & 42.55 & 475.34   & \multirow{2}{*}{95.74} & \multirow{2}{*}{16.46} & 53.19 \\
GPT-5 (context)& 46.81 & 101.70   &  &  & 48.93 \\
\midrule
Random & 16.67 & -  & {16.67} & {-} & -\\
\midrule
Human & 87.23 & 670.91  & {91.49} & 17.50 & \ \  4.26 \\
\bottomrule
\end{tabular}
\caption{Exact match results on {Camlang-CSQA-v0} (ACC: exact match accuracy, Time: average inference time per instance). The last column reports the accuracy drop from English to Camlang.}
\label{tab:csqa-result}
\end{table}

\paragraph{Performance Gap between Camlang and English.}  
All models achieve strong performance on the English version of CommonsenseQA (85–98\% accuracy), confirming that the underlying commonsense reasoning task itself is not the bottleneck. 
The sharp performance drop on Camlang (21–47\%) therefore isolates the difficulty to the metalinguistic reasoning from grammar and dictionary into meaning, which current LLMs fail to operationalise.
Since Camlang requires first parsing questions through explicit grammar and dictionary lookup before reasoning about commonsense relation and knowledge, the drop in accuracy reflects models' inability to perform metalinguistic reasoning in an unknown language. 

\paragraph{Reasoning Models vs. General-Purpose Models.}  
Among LLMs, reasoning systems such as GPT-5 and GPT-o3 achieve the highest accuracies on Camlang (both 46.81\% in the context-only setting), while DeepSeek-R1 reaches 40.43\%. 
In contrast, GPT-4o performs near chance level (21.28\%), only slightly above the random baseline of 16.67\%. 
This suggests that reasoning-oriented models show greater robustness when faced with Camlang's grammar-mediated encoding, though their performance remains far from human.
This is consistent with that Camlang-CSQA indeed requires grammar rule processing before commonsense deduction.
However, whether models actually follow this reasoning path is further examined through human verification in \S~\ref{sec:human_verification}.

\paragraph{Impact of Resource Access Methods.}
Interestingly, we observe that tool use does not always improve the model performance. 
For both GPT-5 and GPT-o3, the context-only configuration outperforms the tools-enabled configuration (46.81 vs.\ 42.55 for GPT-5; 46.81 vs.\ 27.66 for GPT-o3). 
The longer inference times in tool mode do not correlate with higher accuracy, indicating a clear decoupling between time cost and successful rule use.

While CommonsenseQA questions are relatively short multiple-choice instances, the tool setting requires the systems to repeatedly query the grammar book or dictionary, which can add complexity and latency without improving final accuracy.
Another salient factor is that tool-based lookup operates over dictionary stems, while Camlang is morphologically rich and produces many inflected surface forms that may not directly appear in the dictionary. 
In our setup, unless the model can lemmatise first based on the grammar book, direct tool queries may fail to retrieve relevant entries. 
By contrast, in the context-only setting, the full grammar and dictionary are embedded directly into the input, allowing the model to match morphologically related forms through context and thereby avoiding lookup failures.
The result suggests that LLMs struggle to operationalise grammar book and dictionary lookup as part of their reasoning pipeline, even when the API explicitly provides such capabilities.

\paragraph{Human Performance and Learnability Evidence.}
Compared to LLMs, the human participant achieves much higher accuracy on Camlang (87.23\%), only slightly lower than their 91.49\% on English (a drop of 4.26\%). 
This sharp contrast shows that the difficulty for LLMs lies not in the commonsense reasoning component itself, but in the metalinguistic deductive reasoning of Camlang. 
Although solving Camlang questions takes considerably longer for humans (670.9s vs.\ 17.7s), they are able to systematically parse the grammar, consult the dictionary, and integrate the results into reasoning. 
These results are consistent with Camlang being learnable via explicit use of the grammar and dictionary as in \S~\ref{sec:iterative_construction} (cross-annotator translation consistency in the English $\to$ Camlang direction).


\subsection{Human Verification and Analysis on Model Outputs}\label{sec:human_verification}

Although reasoning LLMs such as GPT-5 and GPT-o3 achieve nearly 47\% EM accuracy on Camlang, these numbers alone do not reveal whether correct answers result from genuine reasoning, or from alternative shortcuts such as exploiting surface regularities or relying on English priors.
We therefore turn to human verification of reasoning traces to assess whether correct predictions indeed reflect rule mastery.

\subsubsection{Evaluation Method}
We further verify whether the intermediate reasoning steps of model outputs are correct when their final answers match the gold label. 
In particular, our analysis considers three aspects:
\begin{itemize}
    \item \textbf{Parsing of the Question:} whether the model correctly and completely identified the syntactic structure of the input question (e.g., clause boundaries, argument roles) according to the grammar book.
    \item \textbf{Understanding of the Question Meaning:} whether the model accurately captured the intended semantic interpretation of the question. 
    \item \textbf{Understanding of answer Option Meaning:} whether the model correctly interpreted the meanings of all answer options from the dictionary.
\end{itemize}

We explicitly separate parsing from meaning to distinguish errors in recognising formal structure (syntactic misparsing) from errors in semantic interpretation (misunderstanding word or sentence meaning). 
For instance, a model may arrive at the right meaning despite an incomplete parse by directly mapping key words from the dictionary, or conversely, may parse a structure correctly but still misinterpret semantic roles (see \S~\ref{sec:case_study}, where we present case studies with annotations). 
We do not examine the parsing of options, since most options consist of a single Camlang word translated from one or two English words, leaving little structural ambiguity.

For each aspect, model outputs are categorised into four labels: 
(1) \texttt{correct and complete} (\texttt{Crt+\&Com+}); (2) \texttt{incorrect and complete} (\texttt{Crt-\&Com+}); (3) \texttt{correct and incomplete} (\texttt{Crt+\&Com-}); and (4) \texttt{incorrect and incomplete} (\texttt{Crt-\&Com-}).  
Each instance is first annotated by one annotator and then reviewed by another, and disagreements are subsequently resolved through discussion to reach a final consensus.\footnote{Across 169 instances $times$ 3 aspects (507 annotation in total), only 9 required discussion.}

To provide a better comparison, we report three human verification-based metrics by jointly evaluating all three aspects (question parsing, question meaning, and answer option understanding):
\begin{itemize}
\item \textbf{Strict Human-Verified Accuracy (SHV ACC):} an instance is correct only if it is labelled as \texttt{Crt+\&Com+} in \emph{all} three aspects, reflecting full and faithful rule understanding and application.
\item \textbf{Moderate Human-Verified Accuracy (MHV ACC):} an instance is counted as correct if the question parsing is \texttt{Crt+\&Com+}, while question and option understanding may contain \texttt{incomplete} but not \texttt{incorrect} or \texttt{incorrect and incomplete} labels. This metric requires accurate parsing but tolerates minor omissions in semantic interpretation.
\item \textbf{Lenient Human-Verified Accuracy (LHV ACC):} an instance is counted as correct if question and option understanding contain only \texttt{incomplete} (but not \texttt{incorrect}) labels, regardless of whether parsing is incomplete. 
This captures cases where reasoning succeeds through partial lexical mapping or shallow shortcuts without full grammatical parsing.
\end{itemize}
Note that SHV/MHV/LHV are accuracy-style metrics, defined as the number of instances meeting the respective criteria divided by all test instances (not just EM-correct ones).
Such a tiered comparison allows us to distinguish genuine rule mastery (SHV), parsing-sensitive correctness (MHV), and shortcut-based reasoning (LHV).

\begin{table}[]
    \centering
    \begin{tabular}{l c c c c}
    \toprule
    \textbf{Model} & \textbf{SHV ACC} (\%) & \textbf{MHV ACC} (\%) & \textbf{LHV ACC} (\%) & \textbf{EM ACC} (\%) \\
    \midrule
    4o (tool)     & \ \ 0.00  &\ \ 0.00 &\ \ 0.00  & 21.28 \\
    4o (context) & \ \ 0.00 &  \ \ 0.00&\ \ 0.00 & 21.28 \\
    \midrule
    DS (context) & \ \ 0.00  & \ \ 2.13 &  \ \ 4.26 & 40.43 \\
    \midrule
    o4-mini (tool) & \ \ 0.00 & \ \ 2.13 & \ \ 6.38 &  25.53 \\
    \midrule
    o3 (tool) & \ \ 0.00 & \ \ 2.13  & \ \ 6.38 & 27.66 \\
    o3 (context) &  \ \ 0.00 & \ \ 6.38 &  10.64  & 46.81 \\
    \midrule
    GPT-5 (tool) & \ \ 2.13 & \ \ 2.13 & 17.02 & 42.55 \\
    GPT-5 (context) & \ \ 0.00 & 19.15 & 29.79 & 46.81 \\
    \midrule
    Human & 55.32 & 59.57 & 68.09 & 87.23 \\
    \bottomrule
    \end{tabular}
    \caption{Human verification results on {Camlang-CSQA-v0}. SHV: strict correctness and completeness across all aspects. MHV: requires correct parsing but tolerates incomplete semantic understanding. LHV: allows incomplete but not incorrect interpretations, capturing partial or shortcut reasoning. The last column presents exact-match accuracy (EM ACC) for reference. }
    \label{tab:human_verified_acc}
\end{table}

\subsubsection{Result Discussion}
We report the human verified results in Table~\ref{tab:human_verified_acc}.
Overall, we find a salient gap between humans and LLMs. 
Humans maintain strong performance across all verification criteria, confirming that Camlang is operationally learnable through explicit use of the grammar and dictionary. 
By contrast, LLMs achieve non-trivial EM accuracy scores but almost entirely fail under SHV and MHV, indicating that most correct predictions do not arise from metalinguistic deductive reasoning.
These trends highlight two general conclusions: 
(1) EM substantially overestimates models' reasoning ability, and; 
(2) Camlang effectively exposes the limits of LLMs in integrating explicit formal rules into their reasoning pipeline.  
We next discuss the results in detail.

\paragraph{Sharp decline from EM to SHV.}  
Across all LLMs, SHV accuracy is near to zero, even for GPT-5 (context), which achieves 46.81\% EM. 
This sharp drop suggests that many EM-correct predictions are not supported by faithful grammatical reasoning, but more likely arise from shortcuts such as partial lexical matching or reliance on English priors.
By contrast, humans reach 55.32\% SHV, demonstrating that they can systematically parse questions, reconstruct meanings, and interpret options using explicit rules.  
The contrast reveals that EM alone substantially overstates LLM competence, while SHV directly tests whether Camlang's grammar is truly mastered.  

At the same time, we acknowledge that SHV is deliberately \textit{strict}. LLMs and humans may have partially engaged with the grammar but omitted intermediate explanations they deemed unnecessary, in particular for reasoning LLMs, leading to \texttt{incomplete} labels. Thus, we further analyse MHV and LHV.

\paragraph{Partial Success and Shortcut Reasoning under MHV and LHV.}  
Under MHV, which tolerates incomplete semantics but requires correct parsing, only GPT-5 (context) reaches 19.15\%, while others remain close to zero. 
This confirms that LLMs rarely manage accurate structural parsing.  
LHV further relaxes the criteria by including incomplete-but-not-incorrect interpretations. 
Here, GPT-5 (context) rises to 29.79\% and GPT-5 (tool) to 17.02\%, while o3 (context) reaches 10.64\%. 
These gains suggest that stronger models sometimes align Camlang words with dictionary entries sufficiently to guess the right answer, even without full parsing.  
In other words, their success often comes from shallow alignment rather than genuine metalinguistic deduction.

Moreover, the divergence from EM to LHV makes clear that EM can substantially overestimate grammatical reasoning ability. In particular, EM remains non-trivial for LLMs such as DeepSeek-R1 and o4-mini, while LHV is near zero. This comparison indicates that, for those LLMs, many EM-correct cases arise from shallow heuristics rather than rule-guided parsing.

\paragraph{Human vs.\ LLMs.}  
Humans consistently outperform all LLMs by wide margins under every metric: 55.32\% SHV, 59.57\% MHV, and 68.09\% LHV.  
The drop between EM and SHV indicates humans also resort to shortcuts, such as inferring from lexical cues or skipping minor structures.
However, the high LHV, MHV, and SHV further support that humans can reliably parse the surface syntax and identify the intended hierarchical structure of Camlang sentences.

\subsection{Error Analysis}\label{sec:error_analysis}

We now show the fine-grained error analysis (Table~\ref{tab:hv-distribution}), examining the distribution of correctness and completeness labels across question parsing, question meaning, and option meaning as evaluated in \S~\ref{sec:human_verification}.
Table~\ref{tab:hv-distribution} shows the distribution statistics.

\begin{table}[t]
\centering
\small
\setlength{\tabcolsep}{7pt} 
\begin{tabular}{l c c cc  cc  cc}
\toprule
\multirow{2}{*}{\textbf{Model}} 
& \multirow{2}{*}{\textbf{\# EM}}
& \multirow{2}{*}{\textbf{Label}} 
& \multicolumn{2}{c}{\textbf{Q. Parsing}} 
& \multicolumn{2}{c}{\textbf{Q. Meaning}} 
& \multicolumn{2}{c}{\textbf{O. Meaning}} \\
\cmidrule(lr){4-5} \cmidrule(lr){6-7} \cmidrule(lr){8-9}
 &  &  & \textbf{\#} & \textbf{\% } & \textbf{\#} & \textbf{\% } & \textbf{\#} & \textbf{\% } \\
\midrule
\multirow{4}{*}{4o (tool)} & \multirow{4}{*}{10} 
 & Crt+ \& Com+      & \ \ 0 &\ \ \ \ 0.00 & \ \ 0 &\ \ \ \ 0.00 & \ \ 0 &\ \ \ \ 0.00\\
 & & Crt+ \& Com-\ \     & \ \ 0 &\ \ \ \ 0.00 & \ \ 0 &\ \ \ \ 0.00 & \ \ 0 &\ \ \ \ 0.00\\
 & & Crt-\ \ \& Com+    & \ \ 0 &\ \ \ \ 0.00 & \ \ 0 &\ \ \ \ 0.00 & \ \ 0 &\ \ \ \ 0.00\\
 & & Crt-\ \ \& Com-\ \   & 10 & 100.00 & 10 & 100.00 & 10 & 100.00 \\
\midrule
\multirow{4}{*}{4o (context)} & \multirow{4}{*}{10} 
 & Crt+ \& Com+      & \ \ 0 &\ \ \ \ 0.00 & \ \ 0 &\ \ \ \ 0.00 & \ \ 0 &\ \ \ \ 0.00\\
 & & Crt+ \& Com-\ \     & \ \ 0 &\ \ \ \ 0.00 & \ \ 0 &\ \ \ \ 0.00 & \ \ 0 &\ \ \ \ 0.00\\
 & & Crt-\ \ \& Com+    & \ \ 0 &\ \ \ \ 0.00 & \ \ 0 &\ \ \ \ 0.00 & \ \ 0 &\ \ \ \ 0.00\\
 & & Crt-\ \ \& Com-\ \   & 10 & 100.00 & 10 & 100.00 & 10 & 100.00 \\
\midrule
\multirow{4}{*}{DS (context)} & \multirow{4}{*}{19} 
 & Crt+ \& Com+      & \ \ 4 &\ \ 21.05 & \ \ 6 &\ \ 31.58 & \ \ 9 &\ \ 47.37 \\
 & & Crt+ \& Com-\ \     & \ \ 1 &\ \ \ \ 5.26 & \ \ 0 &\ \ \ \ 0.00 & \ \ 2 &\ \ 10.53 \\
 & & Crt-\ \ \& Com+    & 11 &\ \ 57.89 & 11 &\ \ 57.89 & \ \ 8 &\ \  42.11 \\
 & & Crt-\ \ \& Com-\ \   & \ \ 3 & \ \ 15.79 & \ \  2 &  \ \ 10.53 & \ \ 0 & \ \ \ \ 0.00 \\
\midrule
\multirow{4}{*}{o4-mini (tool)} & \multirow{4}{*}{12} 
 & Crt+ \& Com+  & \ \ 2 &  \ \ 16.67 &  \ \ 1 & \ \ \ \ 8.33  & \ \ 1 & \ \ \ \ 8.33  \\
 & & Crt+ \& Com-\ \   & \ \ 4 & \ \ 33.33  & \ \ 2 & \ \ 16.67 & \ \ 9 & \ \ 75.00 \\ 
 & & Crt-\ \ \& Com+  & \ \ 4 & \ \ 33.33  & \ \ 6 & \ \ 50.00 & \ \ 0 & \ \ \ \ 0.00 \\ 
 & & Crt-\ \ \& Com-\ \   & \ \ 2 &  \ \ 16.67 & \ \ 3 &  \ \ 25.00 &  \ \ 2 & \ \ 16.67 \\
\midrule
\multirow{4}{*}{o3 (tool)} & \multirow{4}{*}{13} 
 & Crt+ \& Com+  & \ \ 5 & \ \ 38.46 &  \ \ 9 & \ \ 69.23 &  \ \ 2 &\ \  15.38   \\
 & & Crt+ \& Com-\ \   &  \ \ 4 & \ \ 30.77 &  \ \ 2 & \ \ 15.38 &  \ \ 5 & \ \ 38.46   \\ 
 & & Crt-\ \ \& Com+  &  \ \ 1  & \ \ \ \ 7.69 & \ \ 2 & \ \ 15.38 &  \ \ 2 & \ \ 15.38 \\
 & & Crt-\ \ \& Com-\ \   & \ \ 3 & \ \ 23.07 &  \ \ 0 & \ \  \ \ 0.00 & \ \ 4 &  \ \ 30.77  \\
\midrule
\multirow{4}{*}{o3 (context)} & \multirow{4}{*}{22} 
 & Crt+ \& Com+    &  \ \ 7 &\ \  31.82 & 15 &\ \  68.18 & \ \ 6 & \ \  27.27   \\
 & & Crt+ \& Com-\ \   &  \ \ 2 &  \ \ \ \ 9.09 &\ \ 1 & \ \ \ \ 4.55 & \ \ 8 & \ \  36.36  \\
 & & Crt-\ \ \& Com+   & \ \ 8 &\ \  36.36 & \ \ 5 &\ \  22.73 & \ \ 8 & \ \  36.36   \\ 
 & & Crt-\ \ \& Com-\ \   &  \ \ 5 &\ \  22.73 & \ \ 1 & \ \ \ \ 4.55 &  \ \ 0 & \ \  \ \ 0.00\\
\midrule
\multirow{4}{*}{GPT-5 (tool)} & \multirow{4}{*}{20} 
 & Crt+ \& Com+  & \ \ 4 & \ \ 20.00 & \ \ 7 & \ \ 35.00 &  \ \ 4 & \ \ 20.00     \\
 & & Crt+ \& Com-\ \    & \ \ 8 &  \ \ 40.00 &  \ \ 6 & \ \ 30.00 &  \ \ 8  & \ \ 40.00  \\ 
 & & Crt-\ \ \& Com+  &   \ \ 2 & \ \ 10.00 &  \ \ 2 &  \ \ 10.00 & \ \ 2 & \ \ 10.00   \\
 & & Crt-\ \ \& Com-\ \  & \ \ 6 & \ \ 30.00 &  \ \ 5 & \ \ 25.00 & \ \ 7 & \ \ 35.00  \\
\midrule
\multirow{4}{*}{GPT-5 (context)} & \multirow{4}{*}{22} 
 & Crt+ \& Com+  &  10 & \ \ 45.45 & 13 & \ \ 59.09 & \ \ 	0  & \ \ \ \ 0.00    \\
 & & Crt+ \& Com-\ \   & \ \  5 & \ \ 22.73 & \ \ 	3& \ \ 13.64	& 20 & \ \ 90.91 \\ 
 & & Crt-\ \ \& Com+   & 	\ \ 4 & \ \ 18.18	& \ \ 3  & \ \ 13.64 & \ \	0 & \ \ \ \ 0.00  \\
 & & Crt-\ \ \& Com-\ \   &  \ \ 3  & \ \ 13.64	& \ \ 3	 & \ \  13.64 & \ \ 2 & \ \  \ \ 9.09\\
 \midrule
\multirow{4}{*}{Human} & \multirow{4}{*}{41} 
 & Crt+ \& Com+  &  34 &\ \ 82.93 & 35 &\ \ 85.37 & 34 &\ \ 82.93 \\
 & & Crt+ \& Com-\ \   &\ \ 4 & \ \ \ \ 9.76 & \ \ 1 & \ \ \ \ 2.44 & \ \ 3 & \ \ \ \ 7.32 \\ 
 & & Crt-\ \ \& Com+   & \ \ 2 & \ \ \ \ 4.88 & \ \ 5 & \ \ 12.20 & \ \ 3 & \ \ \ \ 7.32 \\
 & & Crt-\ \ \& Com-\ \   & \ \ 1 &  \ \ \ \ 2.44 & \ \  0 & \ \ \ \  0.00 &\ \ 1 & \ \ \ \ 2.44 \\
\bottomrule
\end{tabular}
\caption{Distribution of human verification labels for each model, over exact-match-correct instances. Each aspect is broken down into counts (\#) and percentages (\%). 
Note that percentages here are computed within each model's EM-correct instances.
(1) Crt+ \& Com+, (2) Crt+ \& Com-, (3) Crt- \& Com+, and (4) Crt- \& Com- correspond to (1) \texttt{correct and complete}, (2) \texttt{correct and incomplete}, (3) \texttt{incorrect and complete}, and (4) \texttt{incorrect and incomplete}, respectively.}
\label{tab:hv-distribution}
\end{table}

\subsubsection{Analysis across Aspects}
The distribution of verification labels reveals systematic asymmetries across dimensions.

\paragraph{Correct Parsing vs. Correct Meaning.}
Across most LLMs (except o4-mini), parsing is disproportionately weaker than meaning. 
For example, DeepSeek-R1 shows 26.31\% (21.05\% \texttt{Crt+\&Com+} + 5.26\% \texttt{Crt+\&Com-}) of EM-correct cases as in parsing, while 31.58\% in question meaning.
This indicates a tendency to infer plausible semantics directly from lexical cues or commonsense priors without faithfully reconstructing syntactic structure, a form of ``meaning without parsing''. 
In contrast, humans maintain high correctness in both parsing (82.93\% \texttt{Crt+\&Com+} + 9.76\% \texttt{Crt+\&Com-} = 92.69\%) and meaning (85.37 \% \texttt{Crt+\&Com+} + 2.44 \texttt{Crt+\&Com-} = 87.81\%), confirming that they rely on explicit metalinguistic reasoning and interpretation rather than semantic guessing.

\paragraph{Question vs. Option Meaning.}
Another consistent pattern is that some models perform better on interpreting options than on questions. 
For instance, o4-mini achieves 8.33\% \texttt{Crt+\&Com+} and 75\% \texttt{Crt+\&Com-} in option meaning, indicating that despite missing interpretation, its understanding of options is mostly correct.
This contrast suggests that understanding full questions is considerably harder than interpreting isolated options, because questions require reconstructing relationships among multiple words according to the grammar book, rather than mapping single words to dictionary entries.
In other words, option meaning can often be resolved through shallow lexical lookup, whereas question interpretation demands faithful parsing of clause structures and argument roles.
Humans, by contrast, show no such imbalance, with parsing and both meaning dimensions converging at high levels.

\paragraph{Complete vs. Incomplete Reasoning.}
Strong reasoning models such as GPT-5 (context) reach relatively high proportions of \texttt{Crt+\&Com+} in parsing (45.45\%) and question meaning (59.09\%), yet collapse to 0 in option meaning.
Instead, almost all option cases fall into \texttt{Crt+\&Com-}, reflecting reasoning traces where partial lexical alignment is visible but the full interpretation is not explicitly documented.
It suggests that models often omit parts of the reasoning they deem unnecessary, leaving incomplete chains in the written output.

\subsubsection{Analysis across Systems}
As a general-purpose LLM, both tool and context settings of GPT-4o fail almost completely, where all EM-correct cases fall into the \texttt{Crt-\&Com-} category across parsing, question meaning, and option meaning.
This indicates that GPT-4o shows almost no evidence of rule-based reasoning in its traces, and its EM-level success likely comes from random guessing or coincidental alignment with commonsense priors.

Amongst reasoning LLMs, both DS and o4-mini exhibit partial success via shortcuts rather than full parsing.
For DS, option interpretation shows some strength: 11 out of 19 EM-correct cases contain at least partially correct option meanings, yet parsing is mostly incorrect (\texttt{Crt-\&Com+}).
This reflects a ``\textit{meaning without parsing}'' behaviour: the model often guesses semantics from lexical cues and world knowledge, bypassing formal structure.
In contrast, o4-mini's parsing and question meaning remain poor, but 10 (1+9) out of 12 EM-correct cases contain at least partially correct option meanings (mostly \texttt{Crt+\&Com-}), indicating reliance on shallow lexical shortcuts—matching familiar words without reconstructing the question's syntax.


Across both settings, o3 produces partially correct outputs in all three aspects, but rarely aligns them into a complete reasoning step. 
In the {context} setting (\#EM=22), {question meaning} is its relative strength (\texttt{Crt+\&Com+} 15/22 = 68.18\%), while question {parsing} is more scattered (\texttt{Crt+\&Com+} 31.82\%, \texttt{Crt-\&Com+} 36.36\%), indicating that the model often delivers a complete but incorrect parse. 
{Option meaning} is fragmented (\texttt{Crt+\&Com+} 27.27\%, \texttt{Crt+\&Com-} 36.36\%, \texttt{Crt-\&Com+} 36.36\%). 
The {tool} setting (\#EM=13) follows a similar pattern: {question meaning} is relatively strong (\texttt{Crt+\&Com+} 69.23\%), but {option meaning} shows weaker completeness (\texttt{Crt+\&Com+} 15.38\%, \texttt{Crt+\&Com-} 38.46\%). 

GPT-5, particularly in the {context} setting (\#EM=22), achieves the highest totals across parsing (\texttt{Crt+\&Com+} 45.45\%, \texttt{Crt+\&Com-} 22.73\%) and question meaning (\texttt{Crt+\&Com+} 59.09\%). 
However, {option meaning} is almost entirely incomplete (\texttt{Crt+\&Com-} 20/22 = 90.91\%, \texttt{Crt+\&Com+} 0\%), showing that while the model can consistently map options to dictionary entries, it rarely produces a full, verifiable reasoning trace. 
This pattern suggests emerging metalinguistic awareness to a limited extent, where GPT-5 is often incomplete rather than incorrect, but still lacks the systematic, end-to-end grammatical mastery achieved by humans. 
The {tool} setting (\#EM=20) performs weaker overall, with higher proportions of \texttt{Crt-\&Com-} across all aspects (30\% in parsing, 25\% in question meaning, 35\% in option meaning), reflecting failures in handling morphological variation and retrieval. 
Such evidence shows that GPT-5 has the strongest potential, but its reasoning remains fragmented, revealing only partial metalinguistic competence compared to humans' systematic grammar-based reasoning.

Humans consistently rely on structured grammar-based reasoning, in sharp contrast to LLMs' shortcut-based strategies.
Over 80\% of their EM-correct cases are fully correct and complete in parsing, question meaning, and option meaning.

\begin{table}[t]
\centering
\begin{tabular}{p{0.2\linewidth} p{0.7\linewidth}}
\toprule
\textbf{Type} & \textbf{Camlang (English)} \\
\midrule
Question & \textit{nosṇa müś ńer. meni myvá ghöt?} \\
         & (A child wants to survive. What does he need to do?) \\
         \midrule
Option & A. \textit{ṇaw ghöt} (he needs to study) \\
 & B. \textit{dit ghöt} (he needs to fall down) \\
 & C. \textit{ṇat ghöt} (he needs to play) \\
 & D. \textit{wecmylirsi irwéc ghöt} (he needs to ask questions) \\
 & E. \textit{e ghöt} (he needs to eat) \textbf{[correct]} \\
 & F. \textit{nepli cewmyl nak} (none of the above answers) \\
\midrule
Morphemic         & nos = ṇA. müś -m = jer -$\emptyset$. me -nI. mI- n- fa -m = xöt -$\emptyset$. \\
\midrule
Gloss         & child = TOP. live-INF=want-3SG. what-ACC. WH=ACC-do-INF=need-3SG \\
\midrule
Stems & 
\textit{nos}: (n.) child, kid; 
\textit{müś}: (v.) live, survive; 
\textit{jer}: (v.) want; 
\textit{me}: (pro.) Wh; 
\textit{fa}: (v.) do, make, create; 
\textit{xöt}: (v.) must, need, require; 
\textit{ṇaw}: (v.) learn, study (+ACC object); 
\textit{dit}: (n.) fall, fall down; 
\textit{ṇat}: (n.) play; 
\textit{wecmyl}: (n. ORD) question;
\textit{wec}: (v.) ask (someone GEN something ACC); 
\textit{e}: (v.) eat; 
\textit{nepli}: (adj. ORD) true, correct;
\textit{cewmyl}: (n.) answer; 
\textit{e}: (v.) eat; 
\textit{nak}: (v.) do not have, do not exist
\\
\bottomrule
\end{tabular}
\caption{Example Camlang case from Camlang-CSQA-v0, with morphemic segmentation and interlinear gloss.}
\label{tab:case_instance}
\end{table}

\begin{table}[t]
    \centering
    \begin{tabular}{l|ccc}
    \toprule
    \textbf{Models}  & \textbf{Q. Parsing} & \textbf{Q. Meaning} & \textbf{O. Meaning}\\
    \midrule
    4o (tool) & \texttt{Crt-\&Com-} & \texttt{Crt-\&Com-} & \texttt{Crt-\&Com-}\\
    DS (context) & \texttt{Crt-\&Com+} & \texttt{Crt+\&Com+} & \texttt{Crt+\&Com+}\\
    o3 (context) & \texttt{Crt-\&Com-} & \texttt{Crt+\&Com+} & \texttt{Crt-\&Com+}\\
    GPT-5 (tool) & \texttt{Crt-\&Com-} & \texttt{Crt-\&Com-} & \texttt{Crt-\&Com-}\\
    GPT-5 (context) & \texttt{Crt+\&Com-} & \texttt{Crt+\&Com+} & \texttt{Crt+\&Com-}\\
    Human & \texttt{Crt+\&Com+} & \texttt{Crt+\&Com+} & \texttt{Crt+\&Com+}\\

    \bottomrule
    \end{tabular}
    \caption{Human-annotated labels for the same case as in Table~\ref{tab:case_instance}.}
    \label{tab:case_label}
\end{table}

\begin{table}[t]
\centering
\begin{tabular}{p{0.9\linewidth}}
\toprule
\multicolumn{1}{c}{\textbf{GPT-4o (tool)}}\\
\midrule
{\raggedright
To determine the correct answer, let's break down the sentence and options using the resources provided.\\
\vspace{1em}

 \#\#\# Sentence Analysis\\
 \vspace{1em}

 1. "nosṇa müś ńer."\\
  - nosṇa: This word wasn't found in the search results, but based on the structure, it might be a noun or verb.\\
  - müś: This word wasn't found in the search results, but it could be an adjective or adverb.\\
  - ńer: This word wasn't found in the search results, but it might be a verb or noun.\\
 \vspace{1em}

 2. "meni myvá ghöt?"\\
  - meni: This is an interrogative form, similar to "what".\\
  - myvá: This word wasn't found in the search results, but it could be a verb or noun.\\
  - ghöt: This word wasn't found in the search results, but it might be a verb or noun.\\
  \vspace{1em}

 \#\#\# Options Analysis\\
 
 - A: ṇaw ghöt\\
 - B: dit ghöt\\
 - C: ṇat ghöt\\
 - D: wecmylirsi irwéc ghöt\\
 - E: e ghöt\\
 - F: nepli cewmyl nak\\
 \vspace{1em}

 \#\#\# Reasoning\\
 \vspace{1em}

 Given the interrogative "meni," the sentence is likely asking a question about an action or state related to "ghöt." The options seem to provide different actions or states.\\
 
 - "ghöt" appears in several options, suggesting it might be a key verb or noun in the sentence.\\
 - The structure of the sentence and options suggests a focus on an action or state, possibly related to "ghöt."\\
 \vspace{1em}
 
 \#\#\# Conclusion\\
 
 Without specific translations for each word, the best approach is to match the structure and common elements. Since "ghöt" appears in multiple options, it is likely central to the question. The options with "ghöt" are A, B, C, D, and E.\\
 \vspace{1em}
 
 Given the lack of specific translations, the choice is somewhat speculative. However, based on the structure and commonality, Option E: e ghöt seems to be a simple and direct match.\\
 \vspace{1em}
 
 The final answer is E.}\\
\bottomrule
\end{tabular}
\caption{GPT-4o (tool) output of the case example.}
\label{tab:case_4o_context}
\end{table}

\section{Qualitative Analysis and Case Study}\label{sec:case_study}
To better demonstrate our analysis and model behaviour, we present case studies and qualitative analysis, featuring the selected systems.
The case is shown in Table~\ref{tab:case_instance}, with corresponding English translation and human-annotated morphemic and gloss.
We choose this case as its meaning is relatively simple, and it receives correct EM answers from all systems, while the responses differ greatly in the correctness and completeness of analysis.
We present the human-evaluated labels from different aspects in Table~\ref{tab:case_label}.

\begin{table}[t]
\centering
\begin{tabular}{p{0.9\linewidth}}
\toprule
\multicolumn{1}{c}{\textbf{DeepSeek-R1}}\\
\midrule
{\raggedright
The question in Camlang is: "nosṇa müś ńer. meni myvá ghöt?" \\
 - "nosṇa" means "the child" (with topic marker =ṇA). \\
 - "müś" means "to live/survive". \\
 - "ńer" is the 3rd person singular present form of "jer" (to want), with nasal mutation from object proclitic "n=" (for 3sg ACC), so "wants it". \\
 - The first sentence translates to: "The child wants to live." \\
 - The question "meni myvá ghöt?" has: \\
  - "meni" = "what" (ACC). \\
  - "myvá" is not a standard verb; it may be a mishearing or variant of "mavá" (to constitute/make up), but contextually it doesn't fit. \\
  - "ghöt" = "need/must" (from "xöt"). \\
 The question asks: "What does [the child] need to do?" given the context of survival. \\
\vspace{1em}
 Options: \\
 - A: ṇaw ghöt – "need to learn/study" (ṇaw = learn, ORD). Not essential for survival. \\
 - B: dit ghöt – "need to fall" (dit = fall). Illogical for survival. \\
 - C: ṇat ghöt – "need to play" (ṇat = play, ORD). Not critical for survival. \\
 - D: wecmylirsi irwéc ghöt – "need to ask them questions" (wecmylirsi = questions-GEN.PL, irwéc = ask-3pl.ACC). Unrelated to living. \\
 - E: e ghöt – "need to eat" (e = eat, ORD). Eating is fundamental for survival. \\
 - F: nepli cewmyl nak – "true answers do not exist" (nepli = true, cewmyl = answers, nak = not exist). Not an action. \\
\vspace{1em}
 The child needs to eat to survive, making E the correct choice. \\
\vspace{1em}
 The final answer is E.
}\\
\bottomrule
\end{tabular}
\caption{DeepSeek-R1 output of the case example.}
\label{tab:ds_case}
\end{table}

\paragraph{Human Interpretation.}
The question sentence ``\textit{nosṇa müś ńer. meni myvá ghöt?}'' is parsed as follows. 
First, ``\textit{nos=ṇA}'' combines ``\textit{nos}'' (``child'') with the topic marker ``\textit{-ṇA}'', yielding ``\textit{as for the child.}'' 
Next, ``\textit{müś-m= ńer-$\emptyset$}'' forms ``\textit{(he) wants to live}'', where ``\textit{müś}'' (``\textit{live}'') takes the infinitive marker ``\textit{-m=}'', and ``\textit{ńer}'' (``\textit{want}'') carries 3SG agreement. 
The second clause ``\textit{meni myvá ghöt?}'' consists of the $WH$-pronoun ``\textit{meni}'' (``\textit{what-ACC}''), the $WH-$head marking ``\textit{my=}'' plus ``\textit{fa-m=}'' (``\textit{do-INF}'') (orthographically ``\textit{myvá}''), and the modal ``\textit{ghöt-$\emptyset$}'' (``\textit{need-3SG}'') yielding ``\textit{what does he need to do?'}' 
Together, the full sentence means \textit{``A child wants to survive. What does he need to do?''}

All candidate answers (A–E) share the modal ``\textit{ghöt-$\emptyset$}'' (``\textit{need-3SG}''), which attaches to a preceding stem or phrase to form the structure ``\textit{he needs to~X}.'' 
Option A ``\textit{ṇaw ghöt}'' combines ``\textit{ṇaw}'' (``\textit{learn, study}'') with the modal, yielding ``\textit{he needs to study}''.
Option B ``\textit{dit ghöt}'' combines ``\textit{dit}'' (``\textit{fall}'') with the modal, yielding ``\textit{he needs to fall}''. 
Option C ``\textit{ṇat ghöt}'' combines ``\textit{ṇat}'' (``\textit{play}'') with the modal, yielding ``\textit{he needs to play}''.
Option D ``\textit{wecmylirsi irwéc ghöt}'' combines the phrase ``\textit{wecmylirsi irwéc}'' (``\textit{questions-GEN.PL ask-3pl.ACC}'', i.e., ``\textit{ask them questions}'') with the modal, yielding ``\textit{he needs to ask them questions}''.
Option E ``\textit{e ghöt}'' combines ``\textit{e}'' (``\textit{eat}'') with the modal, yielding ``\textit{he needs to eat}''. 
Finally, Option F ``\textit{nepli cewmyl nak}'' is not formed with the modal but instead consists of ``\textit{nepli}'' (``\textit{true}''), ``\textit{cewmyl}'' (``\textit{answers}''), and ``\textit{nak}'' (``\textit{not exist}''), yielding ``\textit{true answers do not exist}''.

\paragraph{GPT-4o.}
We present the output of 4o as shown in Table~\ref{tab:case_4o_context}.
Compared with human interpretation, GPT-4o provides no genuine parsing or semantic interpretation of the question. 
In particular, it fails to identify any morphological markers and produces only vague guesses such as ``\textit{might be a noun or verb}'' showing a complete lack of morphological grounding.
At the level of meaning, 4o does not reconstruct the first clause (``\textit{a child wants to live}'') nor the second clause as a modal question (``\textit{what does he need to do?}''). 
Instead, it merely speculates that ``\textit{meni}'' is interrogative and ``\textit{ghöt}'' is important, without attributing to ``\textit{ghöt}'' its modal sense of ``\textit{need}''.
Also, 4o does not offer any lexical analysis of the options and simply notices that ``\textit{ghöt}'' appears across candidates, choosing the shortest and simplest one. 
The correct answer is therefore obtained by superficial matching rather than by parsing or understanding, effectively resembling random guessing.

\paragraph{DS.}
We present the output of DS as shown in Table~\ref{tab:ds_case}.
For question parsing, DS correctly identifies the topic marker in ``\textit{nos=ṇA}'' and the structure ``\textit{müś-m= ńer-$\emptyset$}'', but it mis-analyses ``\textit{myvá}'' as a possible variant of ``\textit{mavá}'' (``\textit{make up}'') and fails to recognise it as ``\textit{my= fa-m=}'' with the modal ``\textit{ghöt-$\emptyset$}'' (``\textit{need to do X}'').
This leads to incomplete correctness, although the analysis remains largely comprehensive.
Moreover, given the incorrect parsing, DS provides correct interpretations of the question ``\textit{The child wants to live. What does he need to do?}'' and options, which match the human reference exactly and are both correct and complete. 
This provides further evidence that DS relies on contextual \textit{pattern matching} triggered by the survival scenario and the modal ``\textit{ghöt}'', rather than on reconstructing the full syntactic structure.

\begin{table}[t]
\centering
\begin{tabular}{p{0.9\linewidth}}
\toprule
\multicolumn{1}{c}{\textbf{GPT-o3 (context)}}\\
\midrule
{\raggedright
Step-by-step reasoning\\
\vspace{1em}
 1. Lexical meanings \\
 • nos ‘child’ ( Vocab ) \\
 • =ṇA is the topic clitic ( NM1E4, MS6A1 ) \\
 • müś ‘to live, survive’ ( Vocab ) \\
 • jer ‘to want’ → nasal mutation after n-clitic gives ńer ( OP1B3 ) \\
 • meni ‘what-ACC’ ( wh-word with -nI accusative; cf. MS4B1 example meni … ) \\
 • xöt ‘need, must’ → preceded by a nasal wh-object clitic mI=, nasal–fricative coalescence yields ghöt ( OP1D1 ) \\
\vspace{1em}
 2. Literal gloss of the two sentences \\
 nos=ṇA müś n=jer-$\emptyset$.\\
 child.TOP live 3SG.ACC=want-3SG \\
 “As for the child, (he) wants to live.” \\
\vspace{1em}
 meni mI=ghöt ? \\
 what.ACC WH.ACC=need-3SG ? \\
 “What does (he) need?”\\
\vspace{1em}
 3. Required answer: a verb meaning an action that is necessary for a child to keep living. \\
 Among the choices: \\
 A ṇaw ‘study’; B dit ‘fall’; C ṇat ‘play (with toys)’; \\
 D is a long NP, not an action; \\
 E e ‘eat’. \\
\vspace{1em}
 Obviously, eating is the basic requirement for staying alive, matching the context of the question.\\
\vspace{1em}
 4. Surface form ‘e ghöt’ (‘must eat’): \\
 e-m=xöt → phonotactically may also be written e ghöt in casual style, paralleling the model sentence xaxsy e ghötńy in VM8C1.\\
\vspace{1em}
 Therefore option E best answers the question.\\
\vspace{1em}
 The final answer is E
}\\
\bottomrule
\end{tabular}
\caption{GPT-o3 (context) output of the case example.}
\label{tab:o3_context_case}
\end{table}

\paragraph{o3 (context).}
The output of GPT-o3 (context) is shown in Table~\ref{tab:o3_context_case}. 
It correctly analyses ``\textit{nos=ṇA}'' and ``\textit{müś n=jer-$\emptyset$}'', and further provides a literal interlinear gloss. 
However, it mis-parses the second clause as ``\textit{meni mI=ghöt}'' (``\textit{what.ACC WH.ACC=need-3SG}''), overlooking the ``\textit{fa-m=}'' element in ``\textit{myvá}''.
Despite incorrect parsing, o3 captures the intended reading that ``\textit{the child needs (to do) something}'' given the survival context.
Moreover, while covering all options, it incorrectly treats option D (``\textit{wecmylirsi irwéc ghöt}'') as a long NP rather than an action phrase (\textit{ask them questions}''), and motivates \textit{e ghöt}'' via an unsupported alternation, i.e. ``\textit{e-m=xöt} $\Rightarrow$ \textit{e ghöt}'' in ``\textit{casual style}''.

\begin{table}[t]
\centering
\begin{tabular}{p{0.9\linewidth}}
\toprule
\multicolumn{1}{c}{\textbf{GPT-5 (tool)}}\\
\midrule
{\raggedright
Reasoning:\\
 - nosṇa ends with the topic marker =ṇA, so the first sentence sets “nos …” as the sentence topic; that matches how topics are marked in Camlang (e.g., śaṇ=ṇA “today=TOP”) .\\
 - The second sentence is a wh-question with meni “what,” which is exactly how wh-questions are formed in Camlang (wh-expression fronted; no yes-no proclitic) .\\
 - ghöt is the modal verb “need,” and Camlang modals take infinitive complements (not nouns). The grammar illustrates this with e ‘eat’ as the infinitive complement: xaxsy e ghötńy “I need to eat something” and pyjé … ghötly “Do you need to eat …?” .\\
 - Therefore, the natural answer to “What do we need (to do)?” is a verb in the infinitive domain, i.e., e “eat” as the complement of ghöt. The other options are either nouns used as complements to ghöt (ungrammatical for modals) or use nak, the negative copula for adjectival/postpositional predicates, which is irrelevant here .\\
\vspace{1em}
 So, among the choices, only E: e ghöt matches the required “INF + need” structure for modals in Camlang .\\
\vspace{1em}
 The final answer is E.
}\\
\midrule
\multicolumn{1}{c}{\textbf{GPT-5 (context)}}\\
\midrule
{\raggedright
‘nos=ṇA’ = topic “as for the child”; ‘müś’ = “live, survive”; ‘meni … ghöt?’ = “what is needed?” \\
 So the question asks: “For a child to live, what is needed?” \\
 Among the options, the only sensible minimal requirement expressed is “e ghöt” = “(one) needs to eat.”\\
\vspace{1em}
 The final answer is E.
}\\
\bottomrule
\end{tabular}
\caption{GPT-5 (tool) and GPT-5 (context) outputs of the case example.}
\label{tab:gpt5_case}
\end{table}

\paragraph{GPT-5 (tool and context).}
The outputs of GPT-5 (tool) and GPT-5 (context) are shown in Table~\ref{tab:gpt5_case}.
Compared with other systems, the responses of GPT-5 are relatively shorter and simpler, omitting many reasoning steps (although we set both \texttt{verbosity} and \texttt{reasoning effort} as \texttt{high}).
In the tool setting, GPT-5 only notes the topic marker in ``\textit{nos=ṇA}'' and the modal ``\textit{ghöt}'', but it ignores the internal structure of ``\textit{müś-m= ńer-$\emptyset$}'' and fails to analyse ``\textit{myvá}'' as ``\textit{my= fa-m=}''. 
Its paraphrase of the question as ``\textit{What do we need (to do)?}'', omitting the explicit subject and survival context, and it incorrectly rules out other verb options as ungrammatical nouns.
In contrast, the context setting gives a more accurate interpretation.
It correctly identifies ``\textit{nos=ṇA}'' and ``\textit{müś}'' and renders the question as ``\textit{For a child to live, what is needed?}''. Although it still does not parse ``\textit{myvá}'' as \textit{my= fa-m=}'', it provides a correct overall meaning and selects the right option without analysing the distractors.

\section{Conclusion and Future Work}
We presented Camlang, a typologically plausible yet unattested constructed language equipped with a grammar book and dictionary, designed to evaluate whether LLMs can engage in explicit metalinguistic deductive reasoning. 
Through the construction of Camlang-CSQA-v0, we showed that while humans can successfully acquire Camlang and integrate it with commonsense reasoning with existing knowledge, current state-of-the-art LLMs fall short, often relying on shallow alignment rather than systematic grammatical mastery. 

As this is an ongoing project, we plan to expand Camlang and our evaluation framework in several directions.
First, we will scale Camlang to a broader suite of tasks, including translation, grammar questions, and mathematical reasoning, in order to probe different aspects of linguistic and cognitive competence. 
Second, we will extend the human experiments to a larger pool of participants, providing stronger empirical evidence on Camlang's learnability.
Third, we will perform more comprehensive model evaluations.

\section*{Author Contributions}
Fenghua Liu is the core contributor to the Camlang development, Camlang-CSQA-v0 annotation, human verification evaluation, and contributes to the paper writing, in particular for the Camlang language section (\S~\ref{sec:camlang_language}) and case study section (\S~\ref{sec:case_study}).
Yulong Chen proposes the idea, leads the project, conducts the LLM experiments and evaluation, and drafts this manuscript.
Yixuan Liu, Zhujun Jin and Solomon Tsai contribute to the Camlang development, in particular, the refinement process.
Yixuan Liu additionally contributes to the proofreading and revision of this manuscript, and the human participant experiment.
Zhujun Jin additionally contributes to the evaluation in \S~\ref{sec:camlang_analysis}.
Ming Zhong contributes to the discussion.


\section*{Acknowledgments}
We appreciate Theresa Biberauer and Dora Alexopoulou for their suggestions and evaluation during the development of {Camlang}.
We also thank Meiru Zhang, Sana Kidwai, Rami Aly, Leyang Cui, Weiwei Sun, Yufang Hou, Fei Xia, and Bonnie Webber for their discussion of this work.
This work receives financial support from Cambridge Language Sciences and AI@Cam (NRBF.GAAB).
Yulong Chen is further supported by the ERC grant \textsc{AVeriTeC} (GA 865958).


\bibliography{colm2025_conference}
\bibliographystyle{colm2025_conference}

\newpage

\appendix
\section{Typological Features of {Camlang}}\label{sec:wals}
To compare {Camlang} with other known languages, we manually annotate its profile using the features from the World Atlas of Language Structures (WALS)~\citep{haspelmath2005world}. 
The full set of features is presented in~\autoref{tab:camlang_feature}. 
For the fairness of future evaluation and to avoid the possibility that models exploit detailed typological cues for grammar guessing, we do not provide fine-grained feature descriptions.
Also, for simplicity, we do not show features that are not applicable.

\begin{longtable}{@{}l p{0.45\textwidth}  c c}
\caption{WALS-style features of {Camlang}}\label{tab:camlang_feature}\\
\toprule
\textbf{ID} & \textbf{Name} & \textbf{Area} & \textbf{{Camlang}}\\
\midrule
\endfirsthead
\toprule
\textbf{ID} & \textbf{Name} & \textbf{Area} & \textbf{{Camlang}}\\
\midrule
\endhead
\midrule
\multicolumn{4}{r}{\small Continued on next page}\\
\bottomrule
\endfoot

\bottomrule
\endlastfoot

1A & Consonant Inventories & Phonology & 5 \\
2A & Vowel Quality Inventories & Phonology & 3 \\
3A & Consonant-Vowel Ratio & Phonology & 5 \\
4A & Voicing in Plosives and Fricatives & Phonology & 4 \\
5A & Voicing and Gaps in Plosive Systems & Phonology & 2 \\
6A & Uvular Consonants & Phonology & 1 \\
7A & Glottalized Consonants & Phonology & 1 \\
8A & Lateral Consonants & Phonology & 4 \\
9A & The Velar Nasal & Phonology & 1 \\
10A & Vowel Nasalization & Phonology & 2 \\
11A & Front Rounded Vowels & Phonology & 2 \\
12A & Syllable Structure & Phonology & 2 \\
13A & Tone & Phonology & 1 \\
14A & Fixed Stress Locations & Phonology & 1 \\
15A & Weight-Sensitive Stress & Phonology & 8 \\
16A & Weight Factors in Weight-Sensitive Stress Systems & Phonology & 1 \\
17A & Rhythm Types & Phonology & 5 \\
18A & Absence of Common Consonants & Phonology & 1 \\
19A & Presence of Uncommon Consonants & Phonology & 1 \\
20A & Fusion of Selected Inflectional Formatives & Morphology & 1 \\
21A & Exponence of Selected Inflectional Formatives & Morphology & 1 \\
21B & Exponence of Tense-Aspect-Mood Inflection & Morphology & 1 \\
22A & Inflectional Synthesis of the Verb & Morphology & 5 \\
23A & Locus of Marking in the Clause & Morphology & 3 \\
24A & Locus of Marking in Possessive Noun Phrases & Morphology & 3 \\
25A & Locus of Marking: Whole-language Typology & Morphology & 3 \\
25B & non zero marking & Morphology & 2 \\
26A & Prefixing vs. Suffixing in Inflectional Morphology & Morphology & 3 \\
27A & Reduplication & Morphology & 1 \\
28A & Case Syncretism & Morphology & 4 \\
29A & Syncretism in Verbal Person/Number Marking & Morphology & 3 \\
30A & Number of Genders & Nominal Category & 1 \\
31A & Sex-based and Non-sex-based Gender Systems & Nominal Category & 1 \\
32A & Systems of Gender Assignment & Nominal Category & 1 \\
33A & Coding of Nominal Plurality & Nominal Category & 2 \\
34A & Occurrence of Nominal Plurality & Nominal Category & 6 \\
35A & Plurality in Independent Personal Pronouns & Nominal Category & 8 \\
36A & The Associative Plural & Nominal Category & 4 \\
37A & Definite Articles & Nominal Category & 3 \\
38A & Indefinite Articles & Nominal Category & 2 \\
39A & Inclusive/Exclusive Distinction in Independent Pronouns & Nominal Category & 5 \\
40A & Inclusive/Exclusive Distinction in Verbal Inflection & Nominal Category & 5 \\
41A & Distance Contrasts in Demonstratives & Nominal Category & 2 \\
42A & Pronominal and Adnominal Demonstratives & Nominal Category & 1 \\
43A & Third Person Pronouns and Demonstratives & Nominal Category & 1 \\
44A & Gender Distinctions in Independent Personal Pronouns & Nominal Category & 6 \\
45A & Politeness Distinctions in Pronouns & Nominal Category & 1 \\
46A & Indefinite Pronouns & Nominal Category & 2 \\
47A & Intensifiers and Reflexive Pronouns & Nominal Category & 1 \\
48A & Person Marking on Adpositions & Nominal Category & 2 \\
49A & Number of Cases & Nominal Category & 3 \\
50A & Asymmetrical Case-Marking & Nominal Category & 2 \\
51A & Position of Case Affixes & Nominal Category & 1 \\
52A & Comitatives and Instrumentals & Nominal Category & 2 \\
53A & Ordinal Numerals & Nominal Category & 4 \\
54A & Distributive Numerals & Nominal Category & 1 \\
55A & Numeral Classifiers & Nominal Category & 1 \\
56A & Conjunctions and Universal Quantifiers & Nominal Category & 1 \\
57A & Position of Pronominal Possessive Affixes & Nominal Category & 1 \\
58A & Obligatory Possessive Inflection & Nominal Syntax & 1 \\
58B & Number of Possessive Nouns & Nominal Syntax & 1 \\
59A & Possessive Classification & Nominal Syntax & 1 \\
60A & Genitives, Adjectives and Relative Clauses & Nominal Syntax & 6 \\
61A & Adjectives without Nouns & Nominal Syntax & 4 \\
62A & Action Nominal Constructions & Nominal Syntax & 2 \\
63A & Noun Phrase Conjunction & Nominal Syntax & 1 \\
64A & Nominal and Verbal Conjunction & Nominal Syntax & 2 \\
65A & Perfective/Imperfective Aspect & Verbal Categories & 1 \\
66A & The Past Tense & Verbal Categories & 1 \\
67A & The Future Tense & Verbal Categories & 2 \\
68A & The Perfect & Verbal Categories & 3 \\
69A & Position of Tense-Aspect Affixes & Verbal Categories & 2 \\
70A & The Morphological Imperative & Verbal Categories & 2 \\
71A & The Prohibitive & Verbal Categories & 1 \\
72A & Imperative-Hortative Systems & Verbal Categories & 1 \\
73A & The Optative & Verbal Categories & 1 \\
74A & Situational Possibility & Verbal Categories & 2 \\
75A & Epistemic Possibility & Verbal Categories & 3 \\
76A & Overlap between Situational and Epistemic Modal Marking & Verbal Categories & 3 \\
77A & Semantic Distinctions of Evidentiality & Verbal Categories & 2 \\
78A & Coding of Evidentiality & Verbal Categories & 2 \\
79A & Suppletion According to Tense and Aspect & Verbal Categories & 4 \\
79B & Suppletion in Imperatives and Hortatives & Verbal Categories & 5 \\
80A & Verbal Number and Suppletion & Verbal Categories & 1 \\
81A & Order of Subject, Object and Verb & word order & 1 \\
82A & Order of Subject and Verb & word order & 1 \\
83A & Order of Object and Verb & word order & 1 \\
84A & Order of Object, Oblique, and Verb & word order & 3 \\
85A & Order of Adposition and Noun Phrase & word order & 1 \\
86A & Order of Genitive and Noun & word order & 1 \\
87A & Order of Adjective and Noun & word order & 1 \\
88A & Order of Demonstrative and Noun & word order & 1 \\
89A & Order of Numeral and Noun & word order & 1 \\
90A & Order of Relative Clause and Noun & word order & 2 \\
90B & Prenominal relative clauses & word order & 1 \\
91A & Order of Degree Word and Adjective & word order & 1 \\
92A & Position of Polar Question Particles & word order & 5 \\
93A & Position of Interrogative Phrases in Content Questions & word order & 2 \\
94A & Order of Adverbial Subordinator and Clause & word order & 1 \\
95A & Relationship between the Order of Object and Verb and the Order of Adposition and Noun Phrase & word order & 1 \\
96A & Relationship between the Order of Object and Verb and the Order of Relative Clause and Noun & word order & 1 \\
97A & Relationship between the Order of Object and Verb and the Order of Adjective and Noun & word order & 1 \\
98A & Alignment of Case Marking of Full Noun Phrases & simple clauses & 2 \\
99A & Alignment of Case Marking of Pronouns & simple clauses & 2 \\
100A & Alignment of Verbal Person Marking & simple clauses & 2 \\
101A & Expression of Pronominal Subjects & simple clauses & 2 \\
102A & Verbal Person Marking & simple clauses & 5 \\
103A & Third Person Zero of Verbal Person Marking & simple clauses & 4 \\
104A & Order of Person Markers on the Verb & simple clauses & 3 \\
105A & Ditransitive Constructions: The Verb 'Give' & simple clauses & 1 \\
106A & Reciprocal Constructions & simple clauses & 2 \\
107A & Passive Constructions & simple clauses & 1 \\
108A & Antipassive Constructions & simple clauses & 3 \\
108B & Productivity of the Antipassive Construction & simple clauses & 4 \\
109A & Applicative Constructions & simple clauses & 8 \\
109B & Other Roles of Applied Objects & simple clauses & 5 \\
111A & Nonperiphrastic Causative Constructions & simple clauses & 2 \\
112A & Negative Morphemes & simple clauses & 2 \\
113A & Symmetric and Asymmetric Standard Negation & simple clauses & 1 \\
115A & Negative Indefinite Pronouns and Predicate Negation & simple clauses & 1 \\
116A & Polar Questions & simple clauses & 1 \\
117A & Predicative Possession & simple clauses & 2 \\
118A & Predicative Adjectives & simple clauses & 2 \\
119A & Nominal and Locational Predication & simple clauses & 2 \\
120A & Zero Copula for Predicate Nominals & simple clauses & 1 \\
121A & Comparative Constructions & simple clauses & 1 \\
122A & Relativization on Subjects & complex sentences & 4 \\
123A & Relativization on Obliques & complex sentences & 4 \\
124A & 'Want' Complement Subjects & complex sentences & 1 \\
125A & Purpose Clauses & complex sentences & 3 \\
126A & 'When' Clauses & complex sentences & 2 \\
127A & Reason Clauses & complex sentences & 2 \\
128A & Utterance Complement Clauses & complex sentences & 2 \\
143A & Order of Negative Morpheme and Verb & word order & 1 \\
143E & Preverbal Negative Morphemes & word order & 1 \\
143F & Postverbal Negative Morphemes & word order & 4 \\
143G & Minor morphological means of signaling negation & word order & 4 \\
144A & Position of Negative Word With Respect to Subject, Object, and Verb & word order & 7 \\
144B & Position of negative words relative to beginning and end of clause and with respect to adjacency to verb & word order & 3 \\
144L & The Position of Negative Morphemes in SOV Languages & word order & 3 \\
144R & SONegV Order & word order & 1 \\

\end{longtable}

\section{Root Generator}\label{sec:root_generator}

To construct the lexical roots of {Camlang}, we designed a random root generator based on weighted phoneme selection.
Each phonotactic slot (e.g., initial, medial, coda) is associated with a probability distribution over possible phonemes, specified in a frequency table derived from the designed phoneme inventory.
During generation, phonemes are sampled proportionally to these frequencies, and concatenated to form monosyllabic (CVC-type) or bisyllabic (CVCVCV-type) roots.
To ensure naturalistic distribution, duplicates are discarded and sound iconicity is considered (e.g., front vowels for sharp objects, round vowels for round objects).
This procedure yields a pool of candidate roots that respects the intended phonotactic constraints while maintaining variety and plausibility.

\begin{algorithm}[t]
\caption{Random Root Generator for {Camlang}}
\label{alg:rootgen}
\begin{algorithmic}[1]
\State \textbf{Input:} Phoneme inventories $\{P_i\}$ and frequency tables $\{F_i\}$ for each position $i$
\State \textbf{Output:} Set of generated monosyllabic and bisyllabic roots
\Statex

\Function{WeightedRandomChoice}{$P, F$}
    \State Sample a phoneme $p \in P$ with probability proportional to its frequency in $F$
    \State \Return $p$
\EndFunction

\Function{GenerateMonosyllabicRoot}{}
    \State $c_1 \gets$ \Call{WeightedRandomChoice}{$P_1, F_1$}
    \State $c_2 \gets$ \Call{WeightedRandomChoice}{$P_2, F_2$}
    \State $c_3 \gets$ \Call{WeightedRandomChoice}{$P_3, F_3$}
    \State \Return concatenate($c_1, c_2, c_3$)
\EndFunction

\Function{GenerateBisyllabicRoot}{}
    \State $c_1 \gets$ \Call{WeightedRandomChoice}{$P_1, F_1$}
    \State $c_2 \gets$ \Call{WeightedRandomChoice}{$P_2, F_2$}
    \State $c_3 \gets$ \Call{WeightedRandomChoice}{$P_{3a}, F_{3a}$}
    \State $c_4 \gets$ \Call{WeightedRandomChoice}{$P_4, F_4$}
    \State $c_5 \gets$ \Call{WeightedRandomChoice}{$P_5, F_5$}
    \State $c_6 \gets$ \Call{WeightedRandomChoice}{$P_6, F_6$}
    \State \Return concatenate($c_1, c_2, c_3, c_4, c_5, c_6$)
\EndFunction

\Statex
\For{$n = 1$ to $N_{\text{mono}}$}
    \State Generate and store unique \Call{GenerateMonosyllabicRoot}{}
\EndFor
\For{$n = 1$ to $N_{\text{bi}}$}
    \State Generate and store unique \Call{GenerateBisyllabicRoot}{}
\EndFor
\end{algorithmic}
\end{algorithm}

\section{Prompt Details}\label{sec:prompts}
In addition to the grammar book and dictionary as the input, we add a prompt before the question.
For tool-based systems, the prompts are:
\begin{quote}
    \texttt{Developer message:} \texttt{The uploaded files are the grammar book (Grammar.md) and English--Camlang vocabulary (Vocab.xlsx) for the Camlang language. You are asked to use these two resources to understand and answer the question.}\\
    
    \texttt{Prompt message:} \texttt{Below is a multiple-choice question written in Camlang. You can use tools to help you understand the language. You are allowed to generate reasoning steps and an explanation to demonstrate your choice. However, the final line of your response must be exactly in this format, where \{your\_answer\} is the option letter (A, B, C, D, E, or F): The final answer is \{your\_answer\}.}
\end{quote}

For context-only systems, the prompts are:
\begin{quote}    
    \texttt{Prompt message:} \texttt{
    "You are given the Camlang grammar book (Camlang.md) and English-Camlang vocabulary (Vocab.xlsx) for the Camlang language. You are asked to use these two resources to understand and answer the question in Camlang.}\\\\
    \texttt{=== Camlang Grammar ===}\\\\
    \texttt{\{grammar\_text\}}\\\\
    \texttt{=== English-Camlang Vocabulary ===}\\\\
    \texttt{\{vocab\_tsv\}}\\\\
    \texttt{Below is a multiple-choice question written in Camlang. You are allowed to generate reasoning steps and an explanation to demonstrate your choice. However, the final line of your response must be exactly in this format, where \{your\_answer\} is the option letter (A, B, C, D, E, or F): The final answer is \{your\_answer\}.}
\end{quote}
where \texttt{\{grammar\_text\}} and the \texttt{\{vocab\_tsv\}} are the textual grammar book and dictionary, respectively.
\end{document}